\documentclass[journal]{IEEEtran}

\usepackage{cite}
\usepackage{graphicx,subfigure}
\usepackage{amsmath,amsthm,amssymb,bm}
\usepackage{pifont}
\usepackage{multirow}

\usepackage[hidelinks]{hyperref}
\usepackage{subfigure}
\usepackage{algorithm}
\usepackage{algorithmic}

\usepackage{color}
\usepackage{array}
\usepackage{booktabs}

\begin{document}

\title{Dynamics-Adaptive Continual Reinforcement Learning via Progressive Contextualization}

\author{
        Tiantian Zhang, Zichuan Lin, Yuxing Wang, Deheng Ye, Qiang Fu, \\ 
        Wei Yang, Xueqian~Wang, Bin~Liang,  Bo Yuan, and Xiu Li
\thanks{This work was partly supported by the Science and Technology Innovation 2030-Key Project under Grant 2021ZD0201404 and Tencent Rhino-Bird Research Elite Program. ({\em Corresponding author: Bo Yuan and Xiu Li}.)}

\thanks{Tiantian Zhang is with the Department of Automation, Tsinghua University, Beijing 100084, China, and also with the Tencent AI Lab, Shenzhen 518000, China (e-mail: ztt19@mails.tsinghua.edu.cn).}
\thanks{Zichuan Lin, Deheng Ye, Qiang Fu, and Wei Yang are with the Tencent AI Lab, Shenzhen 518000, China (e-mail: zichuanlin@tencent.com; dericye@tencent.com; leonfu@tencent.com; willyang@tencent.com).}
\thanks{Yuxing Wang, Xueqian Wang and Xiu Li are with the Shenzhen International Graduate School, Tsinghua University, Shenzhen 518055, China (e-mail: wyx20@mails.tsinghua.edu.cn; wang.xq@sz.tsinghua.edu.cn; li.xiu@sz.tsinghua.edu.cn).}
\thanks{Bin Liang is with the Department of Automation, Tsinghua University, Beijing 100084, China (e-mail: liangbin@mail.tsinghua.edu.cn).}
\thanks{Bo Yuan is with the Research Institute of Tsinghua University in Shenzhen, Shenzhen 518057, China (e-mail: boyuan@ieee.org).}


}


\maketitle

\begin{abstract}
A key challenge of continual reinforcement learning (CRL) in dynamic environments is to promptly adapt the RL agent's behavior as the environment changes over its lifetime, while minimizing the catastrophic forgetting of  the learned information. 
To address this challenge, in this article, we propose DaCoRL, i.e., dynamics-adaptive continual RL. 
DaCoRL learns a context-conditioned policy using progressive contextualization, which incrementally clusters a stream of stationary tasks in the dynamic environment into a series of contexts and opts for an expandable multihead neural network to approximate the policy. 
Specifically, we define a set of tasks with similar dynamics as an environmental context and formalize context inference as a procedure of online Bayesian infinite Gaussian mixture clustering on environment features, resorting to online Bayesian inference to infer the posterior distribution over contexts. 
Under the assumption of a Chinese restaurant process prior, this technique can accurately classify the current task as a previously seen context or instantiate a new context as needed without relying on any external indicator to signal environmental changes in advance. 
Furthermore, we employ an expandable multihead neural network whose output layer is synchronously expanded with the newly instantiated context, and a knowledge distillation regularization term for retaining the performance on learned tasks.
As a general framework that can be coupled with various deep RL algorithms, DaCoRL features consistent superiority over existing methods in terms of the stability, overall performance and generalization ability, as verified by extensive experiments on several robot navigation and MuJoCo locomotion tasks.
\end{abstract}


\section{Introduction}
\IEEEPARstart{R}{einforcement} learning (RL) \cite{sutton2018reinforcement} is a major learning paradigm in machine learning for sequential decision making tasks. It aims to train a competent policy for an agent that properly maps states to actions to maximize the cumulative reward by interacting with an environment in a trial-and-error manner. 
Traditional RL algorithms, such as Q-learning\cite{1992Technical} and SARSA\cite{1994On}, have been widely studied as tabular methods and successfully applied to Markov decision processes (MDPs) with finite discrete state-action spaces. The emergence of advanced function approximation techniques based on deep neural networks (DNNs) enables RL to have a higher level of understanding of the physical world \cite{li2023survey}, and solve high-dimensional tasks ranging from playing video games \cite{ye2020towards,huang2021tikick,ye2020mastering,ye2020supervised,chen2021heroes,juewumc} to making real-time decisions on continuous robot control tasks 
\cite{francis2020long,bellemare2020autonomous}. 

The progresses of RL have been predominantly focused on learning a single task with the assumption of a stationary\footnote{A stationary environment is an environment whose dynamics normally represented by the reward and state transition functions of the MDP do not change over time.} and fully-explorable environment for sampling observations. 
Nevertheless, in the real-world, environments are often non-stationary and characterized by ever-changing dynamics such as shifts in the terrain or weather conditions, changes of the target position in robot navigation\cite{jaradat2011reinforcement}, and different traffic inflow rates and demand patterns at different times of a day (e.g., peak and off-peak hours) in vehicular traffic signal control\cite{khetarpal2020towards}. 
These scenarios demand competent RL agents that can continually adapt to new environmental dynamics while retaining performance when the previous environment is encountered again.  
Unfortunately, the above requirements are difficult for existing RL methods to fulfill. On the one hand, storing all past experiences may result in a constant growth in memory consumption and computational power. On the other hand, if the size of the replay buffer is limited, the agent may inevitably suffer from the phenomenon known as catastrophic forgetting\cite{mccloskey1989catastrophic,french1999catastrophic,hadsell2020embracing}, incapable of retaining the knowledge and skills learned in previously encountered situations.

\begin{figure}[t]
  \centering
  \setlength{\abovecaptionskip}{-3pt}
  {\includegraphics[width=0.99\linewidth]{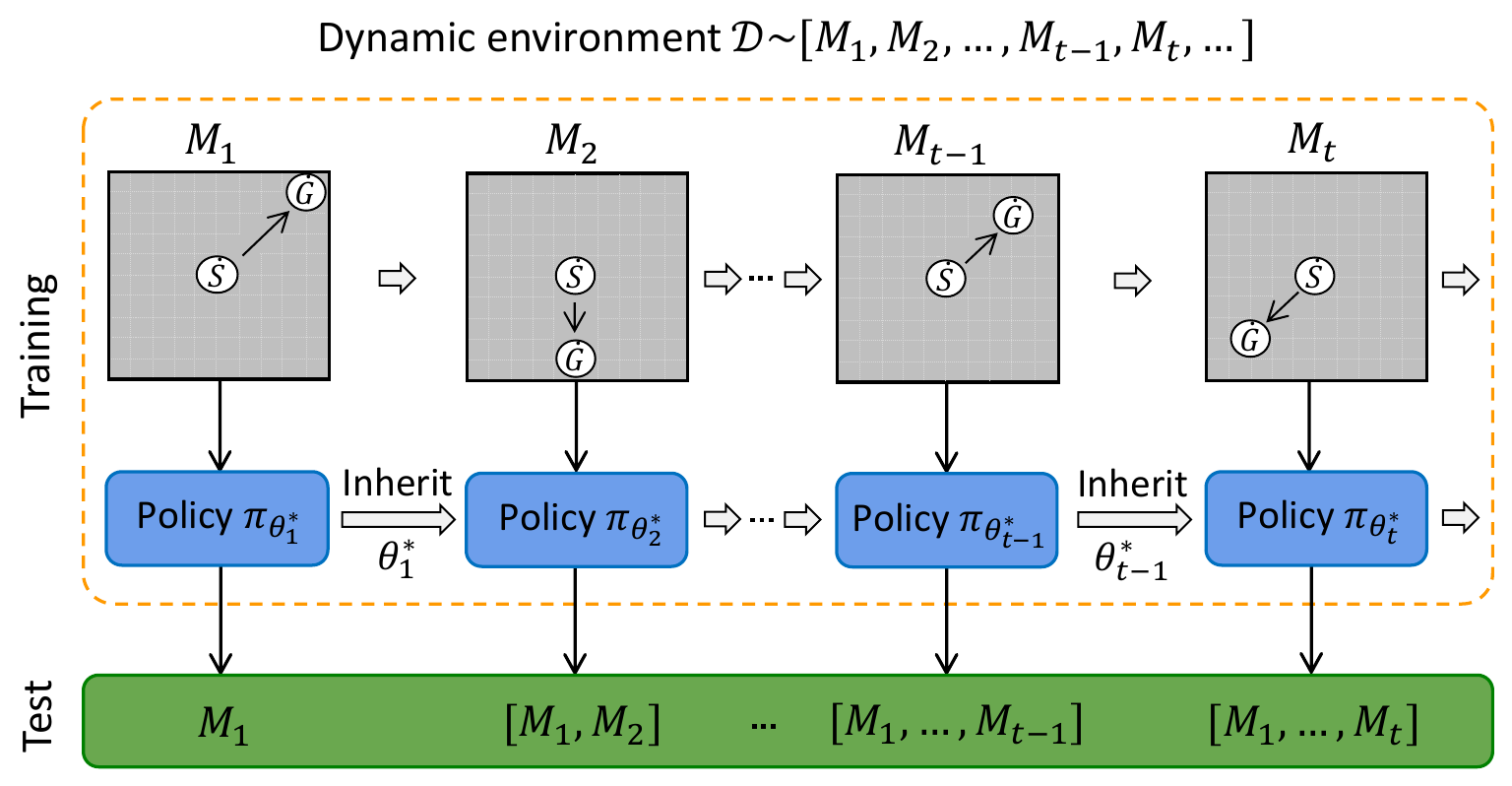}}
  \caption{CRL in dynamic environments. $M_t\in \mathcal{M}, t=1,2,\dots$ denotes the specific MDP/task in time period $t$; $\mathcal{D}$ denotes the dynamic environment over the MDPs space $\mathcal{M}$; $\theta$ are the learning parameters of policy; $\pi_{\theta_t^\ast}$ represents the approximate optimal policy over all learned tasks $[M_1,M_2,\dots,M_t]$.}
  \label{fig:overview_CRL}
\end{figure}

Continual RL (CRL)\cite{kirkpatrick2017overcoming,kessler2020unclear,padakandla2020reinforcement,zhang2022catastrophic} has been investigated as an effective solution for adaptation to dynamic environments and mitigation of catastrophic forgetting. 
In this setting, the dynamic environment can be considered as a stream of stationary tasks on a certain timescale where each task corresponds to the specific environmental dynamics during the associated time period. As shown in Fig. \ref{fig:overview_CRL}, the previously learned policy (e.g., $\pi_{\theta_{t-1}^\ast}$ over $[M_1,M_2,\dots,M_{t-1}]$) is used for the initialization of the new policy (e.g., $\pi_{\theta_t}$ on $M_t$), and it is subsequently updated to fit in the current task during the learning period in a continual fashion, retaining previously learned abilities. 
In other words, CRL is capable of developing proper behaviors for new tasks while keeping the overall performance across all learned tasks. Such features of continual learning are highly desirable for intelligent systems in real-world applications where the environments are subject to consistent changes.

Existing CRL methods assume a decomposition of the original problem into disjoint sub-domains of similar dynamics (also called ``{\em context}") and their boundaries are known in advance. Consequently, previous studies mainly focus on how to construct effective mechanisms to mitigate the catastrophic forgetting among contexts, largely ignoring the challenge of automatic context inference during the learning process. 
For the alleviation of catastrophic interference caused by data distribution drift in the single-task RL, Zhang {\em et al.}\cite{zhang2022catastrophic} employ sequential K-Means clustering to achieve automatic context inference, given the number of contexts ($k$) in advance. 
This method works well because it is feasible to acquire an approximate estimate of the state distribution with sufficient exploration and predetermine the value of $k$ in a single task.
However, significant challenges are expected in dynamic environments, where it is impractical to accurately determine the number of environmental contexts in advance as the changes of environmental dynamics are usually infinite and highly uncertain.
As a result, it is more rational for the agent to infer and instantiate environmental contexts in a fully online and incremental manner during the CRL process.

In this article, we investigate CRL in dynamic environments to achieve continual context inference and necessary adaptation. The ultimate objective is to maximize the overall performance of the agent in the whole environment. 
To this end, we propose a novel dynamics-adaptive continual reinforcement learning scheme (DaCoRL) with progressive contextualization, which incrementally clusters a stream of stationary tasks in dynamic environment into a series of contexts and opts for an expandable multihead neural network to learn a context-conditioned policy.
The progressive contextualization contains two core modules: 
The first one is the incremental context detection procedure for automatically detecting the environmental changes and clustering a set of tasks with similar dynamics into a same context.
The second one is the joint optimization procedure to train the policy online for each unique context using an expandable multihead neural network and a knowledge distillation regularization term. 

To detect the changes of environmental dynamics over time, we introduce the online Bayesian infinite Gaussian mixture model (IGMM)\cite{NIPS1999_97d98119} to cluster environment features in a latent context space, where each cluster corresponds to a separate context. We employ the online Bayesian inference to update the model of contexts in a fully incremental manner, assuming that the prior distribution over the contexts is a Chinese restaurant process (CRP)\cite{pitman2002combinatorial}. 
With this online incremental clustering technique, DaCoRL can incrementally instantiate new contexts according to the concentration parameter\footnote{It is a hyperparameter in CRP denoted by $\alpha$ in this article to control the likelihood of new contexts.} of CRP as needed, without requiring any external information of environmental changes such as the predetermined number of contexts in \cite{zhang2022catastrophic}.
During the joint optimization procedure, we introduce an expandable multihead neural network whose output heads can be adaptively expanded according to the number of instantiated contexts. Compared with the fixed-structured networks, this approach can eliminate unnecessary redundancy of network structure and improve learning efficiency.

The contributions of this article are summarized as follows.
\begin{itemize}
    \item[1)] An incremental context detection strategy is introduced for CRL in dynamic environments. It formalizes context inference as an online Bayesian IGMM clustering procedure, enabling the agent to properly identify changes in environmental dynamics in an online manner without any prior knowledge of contexts.
    \item[2)] A dynamics-adaptive CRL training scheme called DaCoRL is proposed for dynamic environments in continuous spaces. By employing an expandable multihead neural network and a knowledge distillation regularization term, DaCoRL can effectively alleviate the catastrophic forgetting and ensure competitive capacity of RL agents for continual learning in dynamic environments.    
    \item[3)] Extensive experiments on a suite of continuous control tasks ranging from robot navigation to MuJoCo locomotion are conducted to validate the overall superiority of our method over baselines in terms of the stability, overall performance and generalization ability.
\end{itemize}

In the rest part of this article, Section II reviews the related work and Section III introduces CRL problem statement and relevant concepts. In Section IV, the framework of DaCoRL is presented, with details on its mechanism and implementation. Experimental results and analyses on several robot navigation and MuJoCo locomotion tasks are presented in Section V, and this article is concluded in Section VI with some discussions and directions for future work.

\section{Related Work}
Continual learning considers how to learn \textit{a sequence of tasks} while maintaining the performance on previously learned tasks. 
It is conceptually related to incremental learning and online learning as they all assume that tasks or samples are presented in a sequential manner. 
Although incremental learning and continual learning are frequently used interchangeably in the literature, they are not always the same. In some studies\cite{rebuffi2017icarl,rosenfeld2018incremental,hersche2022constrained}, incremental learning is used to describe a learning process where a sequence of incremental tasks are learned in a continual manner --- in this case, continual learning can be referred to as incremental learning. Nevertheless, several other studies on incremental learning \cite{wang2019incremental, wang2021lifelong, alegre2021minimum} concentrate on how to incrementally adjust the previously learned policy to facilitate the fast adaptation to a new task, while not considering the agent performance on old tasks. 
Online learning aims to fit a single model to \textit{a single task} over a sequence of data instances without any adaptation to new tasks or concerns for the mitigation of catastrophic forgetting\cite{bottou1998online,shalev2012online}. 
Another related field is multitask learning\cite{huang2022curriculum,crawshaw2020multi,zhang2021survey}, which tries to train a single model on multiple tasks simultaneously, rather than learning tasks sequentially.

A variety of approaches have been investigated to tackle catastrophic forgetting or interference in the CRL community. They can be classified into three major categories according to how the knowledge of previous tasks is retained and leveraged: replay-based, regularization-based, and expansion-based.
The core idea of replay-based approaches is to store samples of past tasks \cite{isele2018selective,riemer2019learning,rolnick2019experience} or generate pseudo-samples from a generative model \cite{atkinson2021pseudo} to maintain the knowledge about the past in the model. These previous task samples are replayed while learning a new task in the form of either being reused as model inputs for rehearsal \cite{isele2018selective,atkinson2021pseudo} or constraining the optimization of the new task \cite{rolnick2019experience,riemer2019learning}, yielding decent results against catastrophic forgetting. An inherent drawback is the constraint on the memory capacity as the number of tasks grows. Although the generative model can be exempted from this limitation, its own training process can also suffer from forgetting.
Regularization-based approaches protect learned knowledge from forgetting by imposing an extra regularization term on the learning objective, penalizing large updates on important weights\cite{kirkpatrick2017overcoming,kessler2020unclear} or policies\cite{rusu2016policy,traore2019discorl,zhang2022catastrophic} for previous tasks while learning new tasks. This family of solutions is easy to implement and tends to perform well on a small set of tasks, but still faces performance trade-offs on the new and old tasks as the number of tasks increases.
Expansion-based approaches incrementally expand new architectural resources such as the network capacity\cite{rusu2016progressive,mallya2018packnet} or a policy library\cite{alegre2021minimum,wang2021lifelong,wang2022dirichlet} in response to new information. These strategies avoid catastrophic forgetting by protecting all weights for the past tasks from being perturbed by the new information but the knowledge transfer between tasks and the scalability may be limited.

For CRL in dynamic environments, in addition to appropriately adapt the above methods to mitigate catastrophic forgetting, the greatest challenge comes from detecting the changes of environment autonomously. 
For instance, regret minimization algorithms can be applied to MDPs with varying transition probability and reward functions, which assume a finite number and degree of MDP changes and detect them based on sliding windows\cite{auer2008near,gajane2018sliding,ortner2020variational,cheung2020reinforcement} or kernel estimators\cite{domingues2021kernel}. In both cases, the algorithms consider a finite horizon and minimize the regret over this horizon, without any concerns for the mitigation of catastrophic forgetting.
RLCD\cite{da2006dealing} is a model-based approach that estimates the prediction quality of different models and instantiates new ones when none of the existing models performs well. An extension of RLCD replaces the quality measure by the CUSUM-based method to perform change-point detection \cite{hadoux2014sequential}. 
Unlike our method, however, these methods are only applicable to purely discrete settings.

There are also some studies that focus on facilitating fast adaptation to new tasks in continuous dynamic environments. In \cite{canonaco2020model} and \cite{alegre2021minimum}, CUSUM-based change detection mechanisms are combined with well known RL algorithms, such as REINFORCE\cite{williams1992simple} and SAC\cite{haarnoja2018soft,zhou2022revisiting}, to promote adaptation to new tasks.
Nagabandi et al. \cite{nagabandi2018deep} used meta-learning to train a prior over dynamics models that can, when combined with recent data, be rapidly adapted to the new contexts. This method requires an explicit pre-training phase on all tasks, which is not possible under our CRL settings.
LLIRL\cite{wang2021lifelong} is a recently proposed method that employs the EM algorithm, together with a CRP prior on the context distribution, to learn an infinite mixture model to cluster the tasks incrementally over time. In this way, it can selectively retrieve previous experiences of the same clusters to facilitate the adaptation to the new task. 
Similar to \cite{alegre2021minimum}, LLIRL optimizes two separate sets of network parameters to train the behavior policy and parameterize the environment for each instantiated context, respectively, which greatly increases the training and memory cost. Furthermore, the above studies mainly focus on fast adaptation without considering catastrophic interference among in-context tasks and knowledge transfer among between-context tasks, which are critical issues to be addressed in CRL. 

For CRL in dynamic environments, several efforts focus on model-free methods that perform context division directly based on experienced transitions. Context QL\cite{padakandla2020reinforcement} applies an ODCP algorithm to state-reward sequences to detect environmental changes. Afterwards, it either learns a new Q table for the newly detected context, or improves the policy learned if the current context has been previously experienced. This work assumes a preknown pattern and a finite number of changes, since ODCP can only determine whether the context has changed, instead of the specific label. Meanwhile, it is only applicable to RL problems with a small and discrete state-action space due to the Q-tables in use.
CRL-Unsup\cite{lomonaco2020continual} detects distributional shifts for continuous dynamic environments by tracking the difference between the short-term and long-term moving averages of rewards. When changes are detected, it consolidates the memory using EWC\cite{kirkpatrick2017overcoming} to prevent catastrophic forgetting. This method requires storing all policy weights learned before each EWC process is triggered and can be problematic in the case of a positive forward transfer followed by a negative backward transfer.
Recently, IQ\cite{zhang2022catastrophic} shows that performing context division by online clustering and training a multihead neural network with the knowledge distillation technique can alleviate the catastrophic interference caused by data distribution drift in the single-task RL. However, it requires a predetermined number of contexts, which limits its applicability in dynamic environments. 

Our proposed method DaCoRL falls in the combination of regularization-based and expansion-based categories. It performs policy learning using an expandable multihead neural network that has been proved in \cite{d2020sharing} to be helpful to generalize the knowledge and result in a more effective feature extraction and transfer in multitask learning, avoiding the burden of multiple networks training and storing. To lift the restriction of predetermined number of contexts, we introduce an incremental context detection module, which can instantiate contexts incrementally as needed. Furthermore, the knowledge distillation technique can effectively reduce the interference among both between-context and in-context tasks, making it possible to conduct effective continual learning in dynamic environments with a single policy network.

\section{Preliminaries and Notations}
The formulation of the CRL problem in the domain of dynamic environments and the related key concepts are introduced in this section. The notations used in this article are summarized in Appendix \ref{appdix:notations_illustration}.

\subsection{Problem Formulation}
{\em 1) RL in Continuous Spaces:} 
RL is commonly studied following the MDP framework\cite{sutton2018reinforcement}, which is defined as a tuple $M=\langle \mathcal{S},\mathcal{A},\mathcal{P},\mathcal{R},\gamma \rangle$, 
where $\mathcal{S}$ is the set of states; $\mathcal{A}$ is the set of actions; $\mathcal{P}:\mathcal{S}\times\mathcal{A}\times\mathcal{S}\rightarrow[0,1]$ is the transition probability function; $\mathcal{R}:\mathcal{S}\times\mathcal{A}\times\mathcal{S}\rightarrow\mathbb{R}$ is the reward function, and $\gamma\in[0,1]$ is the discount factor. At each time step $t\in\mathbb{N}$, the agent moves from $s_t$ to $s_{t+1}$ with probability $p(s_{t+1}|s_t,a_t)$ after it takes action $a_t$, and receives instant reward $r_t$.

Most RL algorithms rely on the mechanism of policy gradient to handle tasks with continuous state and action spaces\cite{duan2016benchmarking}. In such cases, the policy is defined as a function $\pi:\mathcal{S}\times\mathcal{A}\rightarrow[0,1]$, mapping each state to a probability distribution of actions, with $\sum_{a\in\mathcal{A}}\pi(a|s)=1$,$\forall s\in\mathcal{S}$. If only the state space is continuous, the policy can use the Boltzmann distribution to select discrete actions, while when the action space is also continuous, the Gaussian distribution is commonly used. The goal of policy-based RL is to find an optimal policy $\pi^\ast$ with internal parameter $\theta\in\Theta$ that maximizes the expected long-term discount return
\begin{equation}
    \displaystyle
    J(\theta) = \mathbb{E}_{\tau\thicksim\pi_\theta(\tau)}\big[R(\tau)\big]=\mathbb{E}_{\tau\thicksim\pi_\theta(\tau)}\Bigg[\sum_{t=0}^{\infty}\gamma^tr_t\Bigg]
    \label{eq:goal_of_PG}
\end{equation}
where the expectation is over the complete trajectory $\tau$ generated following $\pi_\theta$ until the end of the agent's lifetime.

In basic policy gradient methods (e.g., REINFORCE\cite{williams1992simple}), the action-selection distribution is usually parameterized by a deep neural network trained by taking the gradient ascent with the partial derivative of the objective (i.e., maximize the excepted return) with respect to the policy parameters. The policy gradient can be approximated expressed as 

\begin{equation}
    \displaystyle
    \begin{split}
        \nabla_\theta J(\theta)&=\nabla_\theta\mathbb{E}_{\tau\thicksim\pi_\theta(\tau)}\Big[R(\tau)\Big]\\
        &=\mathbb{E}_{\tau\thicksim\pi_\theta(\tau)}\Big[R(\tau)\nabla\log\pi_\theta(\tau)\Big] \\
        &\approx\frac{1}{N}\sum_{n=1}^{N}R(\tau^n)\nabla_\theta\log\pi_\theta(\tau^n)
    \end{split}
    \label{eq:gradient_of_PG}
\end{equation}
where $(\tau^1,\tau^2,\dots,\tau^N)$ is a batch of learning trajectories sampled from policy $\pi_\theta$ to estimate the return expectation.

{\em 2) CRL in Dynamic Environments:} Following the convention in \cite{wang2021lifelong}, in this article, we consider the dynamic environment as an infinite sequence of stationary tasks where each task corresponds to the specific environmental dynamics within its time period and the same dynamics may recur more than once across different time periods. 
The time period is assumed to be long enough for the agent to get sufficient experience samples to finish policy learning for the associated task. Suppose that there is a space of MDPs denoted as $\mathcal{M}$, and a dynamic environment $\mathcal{D}$ changing over time in $\mathcal{M}$. The CRL agent interacts with $\mathcal{D}=[M_1,M_2,\dots,M_{t-1},M_t,\dots]$, where each $M_t\in\mathcal{M}$ is a specific MDP/task that is stationary in the $t^{\rm th}$ time period.
In practice, the time period $t$ can be thought of as a task inference period, where $M_t$ is the corresponding task within it. Compared with $M_{t-1}$, $M_t$ is either the same task as $M_{t-1}$ or changes into another different new or previous one. In this process, the identity and actual change point of each task are both {\em unknown} to the agent.

It should be highlighted that, to represent the policies for multiple tasks with a single model, the inputs (observations) must at the very least implicitly contain the discriminative information of the tasks. 
In this setting, the goal of CRL agent in dynamic environments in time period $t$ is to extend the acquired knowledge, accumulated from previously learned tasks $[M_1,M_2,\dots,M_{t-1}]$, to the current task $M_t$, to learn a single policy to achieve the maximum return on all learned tasks $[M_1,M_2,\dots,M_t]$:
\begin{equation}
    \displaystyle
    \theta_t^\ast = \mathop{\arg\max}\limits_{\theta}\sum_{i=1}^{t} J_{M_i}(\theta)
    \label{eq:goal_of_CRL}
\end{equation}
where $J_{M_i}$ is the expected return on task $M_i$. 
Note that the agent is expected to learn a sequence of tasks one by one strategically so that it can retain the previously acquired knowledge when learning new tasks.
In other words, the policy $\pi$ with parameter $\theta_t^\ast$ in \eqref{eq:goal_of_CRL} is an approximate optimal policy over all learned tasks $[M_1,M_2,\dots,M_t]$. In this article, the overall performance of the learned policy across all tasks is the primary metric for judging the performance of CRL agents.

\begin{figure*}[t]
  \centering
  \setlength{\abovecaptionskip}{-5pt}
  {\includegraphics[width=0.995\linewidth]{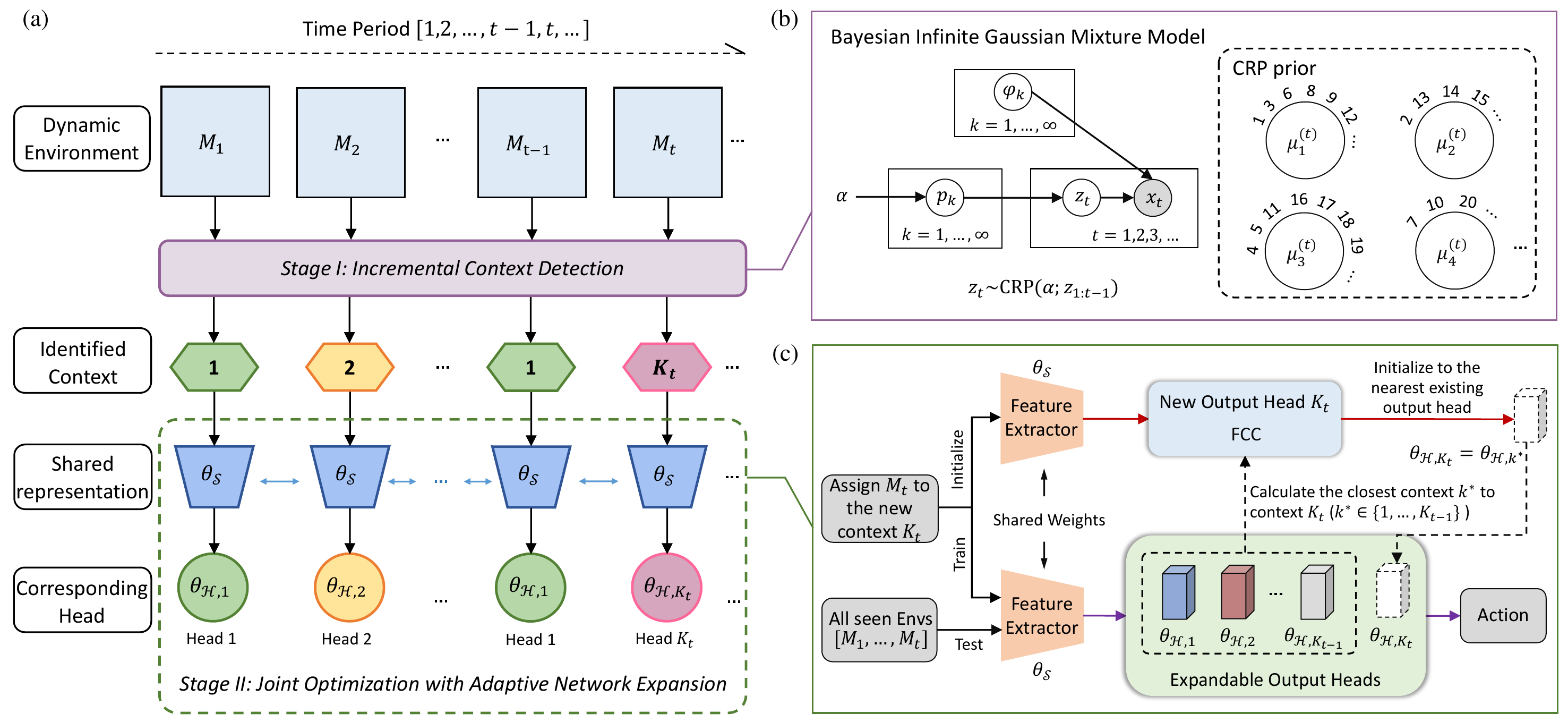}}
  \caption{An overview of DaCoRL in dynamic environments. (a) The general framework of DaCoRL. The dynamic environment is represented by a sequence of stationary tasks $[M_1,M_2,\dots,M_{t-1},M_t,\dots]$ with which the CRL agent interacts sequentially. The incremental context detection module either associates an existing context with the current task (e.g., $M_{t-1}$ belongs to context $1$) or instantiates a new context as needed (e.g., $M_t$ belongs to a new context $K_t$) online. When training sequentially on different tasks, a policy network with shared feature extractor (blue) and a set of expandable output heads corresponding to different contexts is maintained. (b) The Bayesian infinite Gaussian mixture model with the CRP prior for context inference. (c) Multihead neural network expansion. For task $M_t$ assigned to the new context $K_t$, we add a new output head and initialize it to the nearest existing output head, and then train the whole network on $M_t$ while keeping the performance on learned tasks unchanged.}
  \label{fig:DaCoRL_framework}
\end{figure*}

\subsection{Chinese Restaurant Process}
The term CRP\cite{pitman2002combinatorial} arises from an analogy of seating a sequence of customers in a Chinese restaurant with an infinite number of tables (i.e., clusters) and each table has infinite capacity. Each customer sits randomly at an occupied table with probability proportional to the number of current customers at that table, or at an unoccupied table with probability proportional to a hyperparameter $\alpha$ ($\alpha>0$). The conditional probability for the $t^{\rm th}$ customer sitting at the $k^{\rm th}$ table is
\begin{equation}
    \displaystyle
    p(z_t=k|z_{1:t-1}^\ast,\alpha)=\left\{
    \begin{aligned}
    \frac{m_{k}^{(t-1)}}{t-1+\alpha}, \quad & k\leq K_{t-1}\\
    \frac{\alpha}{t-1+\alpha}, \quad & k=K_{t-1}+1
    \end{aligned}
    \right.
    \label{eq:CRP priori}
\end{equation}
where $z_t$ is the latent cluster label of the $t^{\rm th}$ customer; $k$ is the table label and  $z_{1:t-1}^\ast=\{z_1^\ast,z_2^\ast,\dots,z_{t-1}^\ast\}$ represents the table labels assigned to $t-1$ customers, respectively;
$m_{k}^{(t-1)}$ counts the number of assignments to table $k$ up to time $t-1$ and $K_{t-1}$ denotes the number of occupied tables after the $t^{\rm th}$ customer is seated; $\alpha$ serves as the concentration parameter that controls the likelihood of new table.

The CRP can be viewed as equivalent to an IGMM under an assumption that the samples of a CRP are exchangeable\footnote{This means that, under any permutation of samples' ordering, the probability of a particular partitioning result is the same.}
In this article, we employ an IGMM to formulate the distribution of the dynamic environment, that is, we assume that the features of tasks in the dynamic environment present a well-defined Gaussian-like distributions, which is the most common data pattern in practical applications, especially in high-dimensional spaces.
Since tasks with different dynamics appear completely randomly in our problem settings, the prior distribution over this cluster structures can be well described via the CRP, which can provide a flexible structure in which the number of clusters is determined by the observed dynamic environments, and allow IGMM to predict the number of clusters while simultaneously performing model inference.

\section{Proposed Method}
In this section, we give a detailed description of DaCoRL whose overview is shown in Fig.~\ref{fig:DaCoRL_framework}. 
We first introduce the key components including incremental context detection and adaptive network expansion. Then, we present a joint optimization scheme combining the proposed techniques with the canonical policy-based RL to achieve competent CRL in dynamic environments.
For the sake of clarity, we here present an instantiation of DaCoRL using the vanilla policy gradient RL algorithm REINFORCE~\cite{williams1992simple}.

\subsection{Incremental Context Detection}
In dynamic environments, it is important to automatically detect and identify the changes of the environment, which can enable specialized policy learning for different tasks, alleviating the catastrophic forgetting. 
In this article, to characterize the environmental dynamics, we inherit and extend the concept of ``{\em context}" in \cite{zhang2022catastrophic} where a set of tasks with similar dynamics is regarded as the same context. 
Under this setting, the detection of environmental changes can be transformed into a context division procedure in the latent space.

However, compared with the single stationary environment investigated in \cite{zhang2022catastrophic}, context division in dynamic environments brings significantly more challenges. In particular, the changes are usually infinite and highly uncertain and, without prior knowledge, it is hard to determine in advance the number of contexts that need to be instantiated.

To address this issue, we propose an incremental context detection module in DaCoRL, which can instantiate contexts when necessary in an incremental manner without any prior knowledge. 
Specifically, we formalize the task of context detection as an online Bayesian infinite Gaussian mixture clustering procedure. 
To do so, we define a set of environment features to represent the environmental dynamics, and then perform clustering on the features by assuming a CRP prior distribution over the clustering structure. 
In CRP, the instantiation of new contexts is controlled by the concentration parameter $\alpha$: the larger the value of $\alpha$, the more contexts instantiated in general. On the one hand, when $\alpha=0$, only a single context is instantiated during the learning process. On the other hand, when $\alpha\rightarrow\infty$, a new context is instantiated for each task encountered.
The complete procedure of incremental context detection with CRP is described in Algorithm \ref{alg:Incremental Context Detection}, which mostly entails the next two steps.

{\em 1) Environment Feature Construction (lines 1-2):}  
Considering the discriminative task information is implicitly contained in the observation space, we can detect the changes of environmental dynamics by tracking the variation of the observation distribution directly. In practical implementations, we construct the features for a specific task from the collected observations. Specifically, before the start of RL training in each time period, the agent is required to explore the current environment using a random policy
\begin{equation}
    \displaystyle
    \pi_r(a|s)=Uniform(A(s))
    \label{eq:uniform policy}
\end{equation}
where $A(s)$ is the set of available actions in the state $s$ and $Uniform(\cdot)$ is the uniform distribution function. Subsequently, we construct the feature vector ($x_t$) for $M_t$ so that $x_t$ can represent the dynamics of the current environment from the observations $\mathcal{T}_\varepsilon=\{s_0^{i},s_1^{i},s_2^{i},\dots\}_{i=1}^m$, where $m$ is the number of trajectories and $i$ denotes the trajectory index. 

Here, we approximate the set of collected observations as a Gaussian distribution and use its mean vector as the feature of the current environment. Naturally, we can obtain $x_t$ directly from the original observation space 
\begin{equation}
    \displaystyle
    x_t = \frac{1}{N}\sum_{i=1}^{N}s_i
\end{equation}
for the vector inputs, or from the embedding space
\begin{equation}
    \displaystyle
    x_t = \frac{1}{N}\sum_{i=1}^{N}\phi(s_i)
\end{equation}
for the visual inputs, where $N$ is the number of states contained in $\mathcal{T}_\varepsilon$ and $\phi(s)$ can be represented by a random encoder\cite{seo2021state} or a standard GAN model\cite{goodfellow2014generative}.

\begin{algorithm}[t]
    \caption{Incremental Context Detection with CRP}
    \label{alg:Incremental Context Detection}
    \textbf{Input}: The learning task $M_t$ in the $t^{\rm th}$ time period;
    \newline \hspace*{0.95cm} The parameter set $\{\varphi_k^{(t-1)}\}_{k=1}^{K_{t-1}}$ of contexts already
    \newline \hspace*{0.98cm} instantiated before the $t^{\rm th}$ time period.\\
    \textbf{Parameter}: Concentration parameter $\alpha$.\\
    \textbf{Output}: Task-to-context assignment $z_t^\ast$ for $M_t$.
    \newline \hspace*{1.2cm} The parameters $\{\varphi_k^{(t)}\}_{k=1}^{K_{t}}$ of instantiated contexts.

\begin{algorithmic}[1] 
    \STATE Sample $m$ trajectories from $M_t$ using a uniform policy $\pi_r$: $\mathcal{T}_\varepsilon=\{\tau_i\}_{i=1}^m, \tau_i\thicksim\pi_r$.
    \STATE Construct the feature vector $x_t$ of $M_t$ from $\mathcal{T}_\varepsilon$.
    \STATE Initialize $\varphi_{K_{t-1}+1}^{(t-1)}$ of a new potential context as $x_t$.
    \STATE Compute CRP prior $p(z_t=k|z_{1:t-1}^\ast,\alpha)$ for $x_t$.
    \STATE Compute posterior probabilities of task-to-context assignment $p(z_t=k|z_{1:t-1}^\ast,x_t)$.
    \IF{$p(z_t=K_{t-1}+1|z_{1:t-1}^\ast,x_t) > p(z_t=k|z_{1:t-1}^\ast,x_t)$, $\forall k\leq K_{t-1}$}
        \STATE $K_t = K_{t-1}+1$, add $\varphi_{K_{t-1}+1}^{(t-1)}$ to $\{\varphi_k^{(t-1)}\}_{k=1}^{K_{t-1}}$.
    \ELSE
        \STATE $K_t = K_{t-1}$.
    \ENDIF
    \STATE Update parameters $\{\varphi_k^{(t-1)}\}_{k=1}^{K_{t}}$.
    \STATE Calculate $z_t^\ast=\arg\max \limits_{k}{p(x_t|z_t=k,\varphi_k^{(t)})}, \forall k\leq K_{t}$ to obtain the final assignment of $M_t$.
    \RETURN $z_t^\ast$, $\{\varphi_k^{(t)}\}_{k=1}^{K_{t}}$.
\end{algorithmic}
\end{algorithm}

{\em 2) Context Detection via Online Incremental Clustering (lines 3-12):} 
To enable incremental context detection in dynamic environments, based on the constructed environment features, we employ online Bayesian inference to update the context models in a fully online fashion, avoiding the necessity for storing previously seen samples. The key step is to estimate the context parameters $\{\varphi_k\}_{k=1}^{K_t}$, which can build up a mapping between specific tasks and context variables in the latent space. In our implementations, the context model (predictive likelihood function) represents environment features using diagonal Gaussian distributions
\begin{equation}
    \displaystyle
    p(x_t|\varphi_k) = \mathcal{N}(x_t;\mu_k,\sigma^2)
    \label{eq:predictive_function}
\end{equation}
where $\mu_k$ is the mean of the Gaussian used to denote context $k$ and $\sigma^2$ is a constant indicating the variance. Under this setting, the context centroids are just the mean vectors of the Gaussian distributions to be estimated: $\varphi_k=\{\mu_k\}$.

For a sequence of tasks $[M_1,M_2,...,M_{t-1},M_t]$, the first task is assigned to the first context by default.
For $M_t$ in the $t^{\rm th}$ time period, we instantiate a new potential context $K_{t-1}+1$, and initialize the new context model parameter $\varphi_{K_{t-1}+1}^{(t-1)}$ as the feature vector $x_t$, regardless of whether $M_t$ has been encountered before (line 3). After that, the posterior probabilities of the task-to-context assignment over all $K_{t-1}+1$ contexts are estimated (lines 4-5) by
\begin{equation}
    \displaystyle
    \begin{split}
        p(z_t&=k|z_{1:t-1}^\ast,x_t,\varphi_k^{(t-1)},\alpha)\\
        &\varpropto p(x_t|\varphi_k^{(t-1)})p(z_t=k|z_{1:t-1}^\ast,\alpha).
    \end{split}
    \label{eq:posterior1}
\end{equation}

By combining the definition of the predictive likelihood in \eqref{eq:predictive_function} and the CRP prior distribution in \eqref{eq:CRP priori}, the posterior distribution can be rewritten as 
\begin{equation}
    \displaystyle
    \begin{split}
        p(z_t&=k|z_{1:t-1}^\ast,x_t,\varphi_k^{(t-1)},\alpha)\\
        & \varpropto 
        \left\{
            \begin{array}{lll}
            m_{k,t-1}\mathcal{N}(x_t;\mu_k,\sigma^2), & k\leq K_{t-1} \vspace{0.2cm}\\
            \alpha\mathcal{N}(x_t;\mu_k,\sigma^2), & k=K_{t-1}+1
            \end{array}
        \right.
    \end{split}
    \label{eq:posterior2}
\end{equation}
which is abbreviated to $p(z_t=k|z_{1:t-1}^\ast,x_t)$ in subsequent derivations for simplicity.

With the estimated $p(z_t=k|z_{1:t-1}^\ast,x_t)$ for each context, we can determine whether to keep the new context for $M_t$ (lines 6-10). If the posterior probability of the potential context is greater than those of the $K_{t-1}$ existing contexts, this new context is instantiated for $M_t$. 

Next, we employ the EM procedure to update the context parameters in a fully incremental manner. Based on the inferred posterior probabilities, we update context parameters by optimizing the expected log-likelihood
\begin{equation}
    \displaystyle
    \mathcal{L}(x_t|\varphi_k^{(t-1)}) = \mathbb{E}_{M_t}\log p(x_t|z_t=k,\varphi_k^{(t-1)})
    \label{eq:Expectation of likelihood}
\end{equation}
where $M_t\sim p(z_t=k|z_{1:t-1}^\ast,x_t)$. Suppose that all context models with the prior parameters $\varphi_k^{(t-1)}$ have been optimized up to the $(t-1)^{\rm th}$ time period. The estimation of $\{\varphi_k^{(t)}\}_{k=1}^{K_t}$ can be updated based on the gradient (line 11)
\begin{equation}
    \displaystyle
    \varphi_k^{(t)} \gets \varphi_k^{(t-1)} + \eta_{k,t}\nabla_{\varphi_k^{(t-1)}}\mathcal{L}(x_t|\varphi_k^{(t-1)}), \forall k\leq K_t
    \label{eq:paras update}
\end{equation}
where $\eta_{k,t}$ is the learning rate\footnote{Inspired by sequential K-Means clustering adopted in \cite{zhang2022catastrophic}, the updating amplitude of cluster center is inversely proportional to the number of samples contained in the cluster. In our implementations, we set $\eta_{k,t}$ to $\frac{1}{m_{k,t-1}+p(z_t=k|z_{1:t-1}^\ast,x_t)}$, where $m_{k,t-1}$ is the number of tasks assigned to context $k$ up to $(t-1)^{\rm th}$ time period and $p(z_t=k|z_{1:t-1}^\ast,x_t)$ is the posterior probability that $x_t$ is assigned to cluster $k$.} for the $k^{\rm th}$ context in time period $t$. The gradient term $\mathcal{L}(x_t|\varphi_k^{(t-1)})$ can be derived as
$p(z_t=k|z_{1:t-1}^\ast,x_t)\nabla_{\varphi_k}\log p(x_t|z_t=k,\varphi_k^{(t-1)})$.

Based on the updated context parameters, the identity $z_t^\ast$ of $M_t$ can be finally obtained by computing an MAP estimate on the predictive likelihood (line 12), selecting the context model that best fits the current environment.

\subsection{Adaptive Network Expansion}
\label{network_expansion}
To minimize the interference among contexts in CRL, we opt for an expandable neural network in DaCoRL, in which an output head is added synchronously with the newly instantiated context. 
Each output head specializes on a specific context, and the representation layers are shared among different contexts.
This adaptively expandable network structure can parameterize a dedicated policy for each context without any unnecessary redundancy.
In Fig.~\ref{fig:DaCoRL_framework}(a), the set of neural network weights is denoted by $\theta=\{\theta_\mathcal{S},\theta_{\mathcal{H},1},\theta_{\mathcal{H},2},\dots\}$, where $\theta_\mathcal{S}$ is a set of weights for shared representation layers, and $\theta_{\mathcal{H},k}$, $k\in\{1,2,\dots\}$ are the parameters of the $k^{\rm th}$ output head. 

The policy network is initialized as a canonical single-head neural network for the forthcoming learning in the first context. When a new context is instantiated, the neural network is adaptively expanded by adding an output head whose structure is consistent with that of the existing output heads. 
For the newly added output head $K_t$, we propose the following three practical implementations for parameter initialization.
\begin{itemize}                               
    \item[1)] {\em Random:} 
    The weights of the newly added output head are initialized to random values. In this case, the policy of the newly instantiated context inherits the representation module of the learned policies, while its output layer is trained from scratch.
    
    \item[2)] {\em Random-previous Head:} 
    It randomly selects one of the trained output heads and then initializes the weights of the newly added output head to those of the selected one, i.e., $\theta_{\mathcal{H},K_t}=\theta_{\mathcal{H},k}$, where $k=Uniform(\{1,2,\dots,K_{t-1}\})$. This method enables the new policy to inherit knowledge from the learned context. However, it is likely that the cloned policy may hinder the CRL agent's ability to properly explore the task space corresponding to the current context, especially when there are significant differences between the two contexts.

    \item[3)] {\em Nearest-previous Head:} 
    It initializes the newly added output head to a specific trained one whose associated context is nearest to the current instantiated context in the latent context space, i.e., $\theta_{\mathcal{H},K_t}=\theta_{\mathcal{H},k^\ast}$, where $k^\ast=\mathop{\arg\min}_{k}dist(\varphi_k,\varphi_{K_t})$, $k\in\{1,2,\dots,K_{t-1}\}$ and $dist(\varphi_k,\varphi_{K_t})$ denotes the distance between 
    contexts $k$ and $K_t$ in the latent context space. 
\end{itemize}

Intuitively, the third implementation is most likely to encourage forward transfer. The experimental results and analysis in Section \ref{analysis} provide further elaboration on this point.

\subsection{Joint Optimization Scheme}
In the policy optimization stage, we integrate the above components with policy-based RL algorithms to achieve efficient CRL in dynamic environments.
Furthermore, we employ the knowledge distillation technique \cite{zhang2022catastrophic} to mitigate the catastrophic interference caused by distribution drifts among contexts as well as the minor interference due to the differences among tasks within the same context.
Taking REINFORCE as the underlying policy-based RL algorithm as an example, the optimization objective of DaCoRL is derived as follows.

First of all, we rewrite the original loss function of REINFORCE in \eqref{eq:goal_of_PG} with the context label variable $z_t^\ast$ as
\begin{equation}
    \displaystyle
    \mathcal{L}_{ori}(\theta_{z_t^\ast}) = \mathbb{E}_{\tau\thicksim\pi_{\theta_{z_t^\ast}}(\tau)}\big[R(\tau)\big]
    \label{eq:loss_ori}
\end{equation}
where $\theta_{z_t^\ast}=\{\theta_\mathcal{S}, \theta_{\mathcal{H},z_t^\ast}\}$ and $\pi_{\theta_{z_t^\ast}}$ is the policy corresponding to the context associated with the current task $M_t$.

For possible interference among tasks, we adopt the knowledge distillation technique to construct a regularization term in the probability distribution estimation of actions, to preserve the previously learned policies.
Specifically, we regard the policy network from in the last time period as the teacher network, expressed as $\pi_{\theta^-}$, and the current policy network to be trained as the student network, denoted by $\pi_{\theta}$. Then, Kullback-Leibler divergence is used to constrain the difference between this two networks' policies.
Thus, the distillation loss of the output head for context $k$ is defined as  
\begin{equation}
    \displaystyle
    \mathcal{L}_{\mathcal{D}_k}(\theta_k) = \mathbb{E}_{\tau\thicksim\pi_{\theta_{z_t^\ast}}(\tau)}\Big[KL\big[\pi_{\theta_k}(\cdot|s), \pi_{\theta_k^-}(\cdot|s)\big]\Big]
    \label{eq:loss_distill_}
\end{equation}
where $\theta_k=\{\theta_\mathcal{S}, \theta_{\mathcal{H},k}\}$.
In the $t^{\rm th}$ time period, considering all $K_t$ output heads in the policy network, the distillation loss term sums up as
\begin{equation}
    \displaystyle
    \mathcal{L}_{\mathcal{D}}(\theta) = \sum_{k=1}^{K_t} \mathcal{L}_{\mathcal{D}_k}(\theta_k) 
    \label{eq:loss_distill}
\end{equation}
where $\theta=\{\theta_\mathcal{S}, \{\theta_{\mathcal{H},k}\}_{k=1}^{K_t}\}$ denotes the set of all weights of the policy network. 

Finally, to optimize a policy network that can guide the agent to make proper decisions in dynamic environments without being adversely affected by catastrophic forgetting, we combine \eqref{eq:loss_ori} and \eqref{eq:loss_distill} to form a joint optimization scheme. Namely, we solve the CRL problem in dynamic environments by the following optimization objective
\begin{equation}
    \displaystyle
    \begin{array}{lll}
        \max_{\theta_\mathcal{S}, \theta_\mathcal{H}} \mathcal{L}_{ori}(\theta_{z_t^\ast}) - \lambda\mathcal{L}_{\mathcal{D}}(\theta) \vspace{0.2cm}\\
        \theta_{z_t^\ast}=\{\theta_\mathcal{S}, \theta_{\mathcal{H},z_t^\ast}\} \vspace{0.2cm}\\
        \theta=\{\theta_\mathcal{S}, \{\theta_{\mathcal{H},k}\}_{k=1}^{K_t}\}
    \end{array}
    \label{eq:DaCoRL_optimization}
\end{equation}
where $\lambda\in[0,1]$ is a coefficient to control the tradeoff between learning the new policy and preserving the learned policies.

\begin{algorithm}[t]
    \caption{DaCoRL}
    \label{alg:DaCoRL}
    \textbf{Input}: Dynamic environment $\mathcal{D}=[M_1,M_2,\dots,M_T]$.
    \newline \hspace*{0.90 cm} Single-head policy network $\pi_{\theta}$ with random weights 
    \newline \hspace*{0.90 cm} $\theta=\{\theta_{\mathcal{S}}^{(0)}, \theta_{\mathcal{H},1}^{(0)}\}$.\\
    \textbf{Parameter}: Concentration parameter $\alpha$; Learning rate $\beta$;
    \newline \hspace*{1.71cm} Distillation regularization coefficient $\lambda$.\\
    \textbf{Output}: Parameter set $\{\varphi_k\}_{k=1}^{K_T}$ of instantiated contexts;
    \newline \hspace*{1.2cm} Approximate optimal policy parameters $\theta^{\ast}$ for $\mathcal{D}$.
    
\begin{algorithmic}[1] 
    \FOR{each time period $t\in \{1,2,\dots,T\}$}
        \IF{$t=1$}
            \STATE Sample randomly and construct the feature vector $x_1$ for the task $M_1$.
            \STATE Instantiate $M_1$ as the first context: $\varphi_1^{(1)}=x_1$, $z_1^\ast=1$.\\
            \STATE Set $K_1=1$, $m_1^{(1)}=1$ in the CRP model.
            \STATE Update the policy network from scratch using the canonical policy gradient method to obtain $\theta^{\ast}$:
            \vspace{0.1cm}
            \newline \hspace*{1.8cm} $\theta \gets \theta+\beta\nabla_{\theta}\mathcal{L}_{ori}$.
        \ELSE
            \STATE Infer the task-to-context assignment $z_t^\ast$ for $M_t$ using incremental context detection with CRP:\\
            \vspace{0.15cm}
            $z_t^\ast, \{\varphi_k^{(t)}\}_{k=1}^{K_t} \leftarrow{\mbox{Algorithm \ref{alg:Incremental Context Detection}}(M_t,\{\varphi_k^{(t-1)}\}_{k=1}^{K_{t-1}}, \alpha)}$.\\
            \vspace{0.15cm}
            \STATE Update the CRP model according to $z_t^\ast$: 
            \vspace{0.2cm}
            \newline \hspace*{0.3cm} $m_k^{(t)}=m_k^{(t-1)}, \forall k\leq K_t$; \ $m_{z_t^\ast}^{(t)}=m_{z_t^\ast}^{(t-1)}+1$.
            \vspace{0.2cm}
            \IF{$z_t^\ast=K_{t-1}+1$}
                \STATE Expand $\pi_{\theta}$ by adding an output head initialized with the nearest-previous head, add $\theta_{\mathcal{H},K_t}$ to $\theta$.
            \ENDIF
            \STATE Set $\pi_{\theta^-}=\pi_{\theta}$ for distillation.
            \STATE Update the policy network using \eqref{eq:DaCoRL_optimization} to obtain $\theta^\ast$ :
            \vspace{0.1cm}
            \newline \hspace*{1.2cm} $\theta \gets \theta+\beta\nabla_{\theta}(\mathcal{L}_{ori}-\lambda\mathcal{L_{\mathcal{D}}})$.
        \ENDIF
    \ENDFOR
    \RETURN $\{\varphi_k\}_{k=1}^{K_T}$, $\theta^{\ast}$.
\end{algorithmic}
\end{algorithm}

The complete procedure of DaCoRL is summarized in Algorithm \ref{alg:DaCoRL} where the agent interacts with a dynamic environment $\mathcal{D}=[M_1,M_2,\dots,M_T]$. 
In the first time period $t=1$, the task $M_1$ is instantiated as the first context with parameter $\varphi_1^{(1)}$ (line 4). We initialize the CRP model with $K_1=1$, $m_1^{(1)}=1$ (line 5) and train the single-head policy network from scratch using the canonical policy gradient method (line 6).
In the $t^{\rm th}$ ($t\geq 2$) time period, we first apply the incremental context detection module to identify the context to which the current task belongs (line 8), and update the CRP model based on the identity of $M_t$ for future use (line 9). 
Then, we employ an expandable multihead neural network for policy optimization. Specifically, when a new context is instantiated, the policy network is synchronously expanded by adding an output head initialized with the nearest-previous head (lines 10-12). 
Next, the knowledge distillation regularization term is integrated into the loss of the original RL algorithm to reduce the catastrophic forgetting of learned tasks, and the parameters of the policy network are updated till convergence (lines 13-14). 
Finally, $K_T$ contexts with parameters $\{\varphi_k\}_{k=1}^{K_T}$ and the optimal policy $\pi_{\theta^\ast}$ are obtained for all learned tasks.

\section{Experiments and Evaluations}
In this section, we conduct comprehensive experiments on several continuous control tasks from robot navigation to MuJoCo locomotion to demonstrate the effectiveness of our method. 
We design a variety of sequential learning tasks with diverse changes in the underlying dynamics. 
These problem settings are expected to be representative of the dynamic environments that RL agents may encounter in real-world scenarios. 
The following are the overarching questions that we aim to answer from our experiments and analysis.

\begin{itemize}                               
    \item[Q1] Does DaCoRL successfully achieve better continual reinforcement learning in various dynamic environments compared with existing methods?
    
    \item[Q2] How does the initialization strategy of the newly added output head affect the performance of DaCoRL? 
    
    \item[Q3] How does the number of instantiated contexts in the latent space affect the performance of DaCoRL?
    
    \item[Q4] Can DaCoRL achieve a positive forward transfer during the learning process?
    
    \item[Q5] How is DaCoRL's generalization ability to previously unseen tasks?
\end{itemize}

\subsection{Datasets}
{\em 1) Robot Navigation\cite{wang2019incremental,wang2021lifelong}:} It contains three types of dynamic environments with parametric variation across tasks. Types I, II, and III indicate that the dynamic environments are created in terms of the goal position (changes in the reward function), the puddles positions (changes in the state transition function), and both the goal and puddles positions (changes in both above functions), respectively. 
In each task, the agent needs to move to a goal position within a unit square. The state consists of the agent's current 2-D position, and the action corresponds to the 2-D velocity commands in the range of $[-0.1,0.1]$. The reward is equal to the negative squared distance to the goal position minus a small control cost that is proportional to the action's scale. Each learning episode always starts from a given position and terminates when the agent is within $0.01$ from the goal or when the episode length is greater than $100$. We choose these commonly used domains as they are well-understood, suitable for highlighting the mechanism and verifying the effectiveness of our method in a straightforward manner.

{\em 2) MuJoCo Locomotion\cite{wang2021lifelong}:} It contains three locomotion tasks with parametric variation and growing dimensions of state-action spaces.
These tasks require a one-legged hopper, a planar cheetah or a 3-D quadruped ant robot to run at a particular velocity along the positive $x$-direction. The reward is an alive bonus plus a regular part that is negatively correlated with the absolute value between the current velocity of the agent and a preset target velocity. 
The dynamic environment is designed to apply parametric variations in the target velocity within a range: $[0.0, 1.0]$ for Hopper, $[0.0, 2.0]$ for HalfCheetah, and $[0.0, 0.5]$ for Ant. Each learning episode always starts from a given physical status of the agent and terminates when the agent falls down or when the episode length is greater than $100$. We choose these domains to further evaluate the efficiency of our method on more sophisticated domains. 

In this article, we follow the multitask learning benchmark construction described in \cite{yu2020meta}, and extend the original observation space for each task by adding the associated parametric variations, so as to make the observation input discriminative among tasks and make it sensible to represent policies for sequential tasks with a single model\footnote{This is a prerequisite for effective multitask learning with a single model. Serious conflicts in model training will occur when the same observation input corresponds to tasks with different objectives.}. Without loss of generality, we here insert the parametric variations\footnote{The positions of goal/puddles/both of them for Type I/Type II/Type III of navigation tasks and the target velocity for MuJoCo locomotion tasks.} into random dimensions of the initial observation space.
In our experiments, we uniformly generate four Gaussian clusters (i.e., expected contexts) for each dynamic environment in its parametric variation space. Each of the first three clusters consists of 12 tasks and the fourth cluster contains 14 tasks. We sequentially arrange these $T=50$ tasks in a random order, resulting in a dynamic environment $\mathcal{D}=[M_1,M_2,\dots,M_T]$.
For DaCoRL (Oracle), the supervised version of DaCoRL mentioned in the next section, we make available the task-to-context assignments based on the results of cluster generation. By contrast, the correspondence between tasks and contexts is unknown and needs to be identified by DaCoRL itself.

\subsection{Baselines}
We evaluate our method in comparison to the following four state-of-the-art baseline methods and one supervised version of our proposed DaCoRL in dynamic environments. 

\begin{itemize}
    \item[1)] {\em Naive:} It refers to canonical RL methods (e.g., REINFORCE\cite{williams1992simple}, PPO\cite{schulman2017proximal,wu2021coordinated}) that simply train a policy model during the learning process, without paying attention to any possible environmental changes or forgetting.
    
    \item[2)] {\em CRLUnsup\cite{lomonaco2020continual}:} It detects environmental changes by observing the ability of the agent to perform the task. When the changes are detected, it triggers the memory consolidation procedure employing EWC to preserve previously learned policies from catastrophic forgetting. 
    
    \item[3)] {\em CDKD:} It is adapted from the IQ \cite{zhang2022catastrophic} framework that can alleviate catastrophic interference for value-based RL in stationary environments. In this article, we extend IQ to policy-based RL with context division and knowledge distillation (renamed as CDKD), and set the true number of contexts in advance so that it can conduct continual learning in dynamic environments.
    
    \item[4)] {\em LLIRL\cite{wang2021lifelong}:} It is a recent method that focuses on fast adaptation in non-stationary environments, which maintains a library that contains an infinite mixture of parameterized environment models for task clustering and optimizes the learning parameters based on the knowledge accumulated by the previously learned similar tasks to maximize return in the current environment, {\em without} considering the forgetting of previously learned policies.
    
    \item[5)] {\em DaCoRL (Oracle):} This approach can be regarded as a supervised version of DaCoRL, in which the agent is informed in each time period of the specific task-to-context mapping (i.e., context identities are made available).
\end{itemize}

To handle continuous control tasks, we use policy search with nonlinear function approximation in our experiments.
For the navigation tasks, we perform gradient updates using the vanilla policy gradient RL algorithm REINFORCE. In addition, we employ PPO as the base RL algorithm to learn in the more challenging MuJoCo locomotion tasks.

\subsection{Implementation}
We adopt a similar network architecture for all tasks. The policy of DaCoRL is approximated by a feed-forward neural network that contains a fully connected hidden layer (with $200$ units) used as the feature extractor and an expandable output head module used as the action distribution predictor, where each output head consists of a $200$-unit fully connected hidden layer and a fully connected output layer. The hidden layers are connected by ReLU nonlinearity, following the network configuration for these tasks in \cite{wang2021lifelong}.
For a fair comparison, the network architecture of CDKD is set to the same as that of DaCoRL and the number of contexts predetermined for CDKD is also set to be consistent with that automatically detected by DaCoRL. 
For Naive, CRLUnsup, and LLIRL, each policy network consists of two $200$-unit hidden layers connected by ReLU nonlinearity and a fully connected output layer.

During the learning process, we train the model for $1$k policy iterations on each task, and evaluate the policy performance by testing the current policy on all tasks in the dynamic environment every $100$ iterations. All results reported are the average performance over five independent runs.

\subsection{Evaluation Metrics}
Following the convention in \cite{wang2019incremental,wang2021lifelong}, we define two metrics to systematically evaluate our method.
The first one is the average return over a batch of test episodes on all tasks in the dynamic environment, which is used to evaluate the overall performance of the model on all tasks after a fixed number of policy iterations during training in real time. It is defined as
\begin{equation}
    \displaystyle
    \mathcal{R}_{ave} = \frac{1}{Tm}\sum_{i=1}^T\sum_{j=1}^m R(\tau_{ij})
    \label{eq:R_test}
\end{equation}
where $T$ is the number of tasks in the dynamic environment; $m$ is the number of episodes tested for each task; $R(\tau_{ij})$ is the cumulative reward obtained in the $j^{\rm th}$ test episode of task $i$. 
The other is the average return over all test episodes, which evaluates the average performance of the model over the entire training process. It is defined as 
\begin{equation}
    \displaystyle
    \bar{\mathcal{R}}_{ave} = \frac{1}{J}\sum_{i=1}^J \mathcal{R}_{ave}^{(i)}
    \label{eq:average_R_test}
\end{equation}
where $J$ is the number of $\mathcal{R}_{ave}$ evaluations in the learning process, and $\mathcal{R}_{ave}^{(i)}$ is $\mathcal{R}_{ave}$ in the $i^{\rm th}$ evaluation.

{\em Remark:} For task $M$, the test procedure of DaCoRL is conducted in two steps: 
1) Policy selection. It firstly constructs the feature vector $x$ following the procedure in lines 1-2 of Algorithm \ref{alg:Incremental Context Detection}, and then determines the identity $z^\ast$ by computing an MAP estimate on the predictive likelihood of $K_T$ contexts. The policy corresponding to the output head $z^\ast$ is the final policy selected for $M$.
2) Policy execution. The agent applies the selected policy on $M$ to evaluate its performance in terms of the cumulative reward in each episode.  

\begin{figure*}[t]
    \centering
    \setlength{\abovecaptionskip}{3pt}
    \subfigure[\label{fig:typeI}]
        {\includegraphics[width=0.27\linewidth]{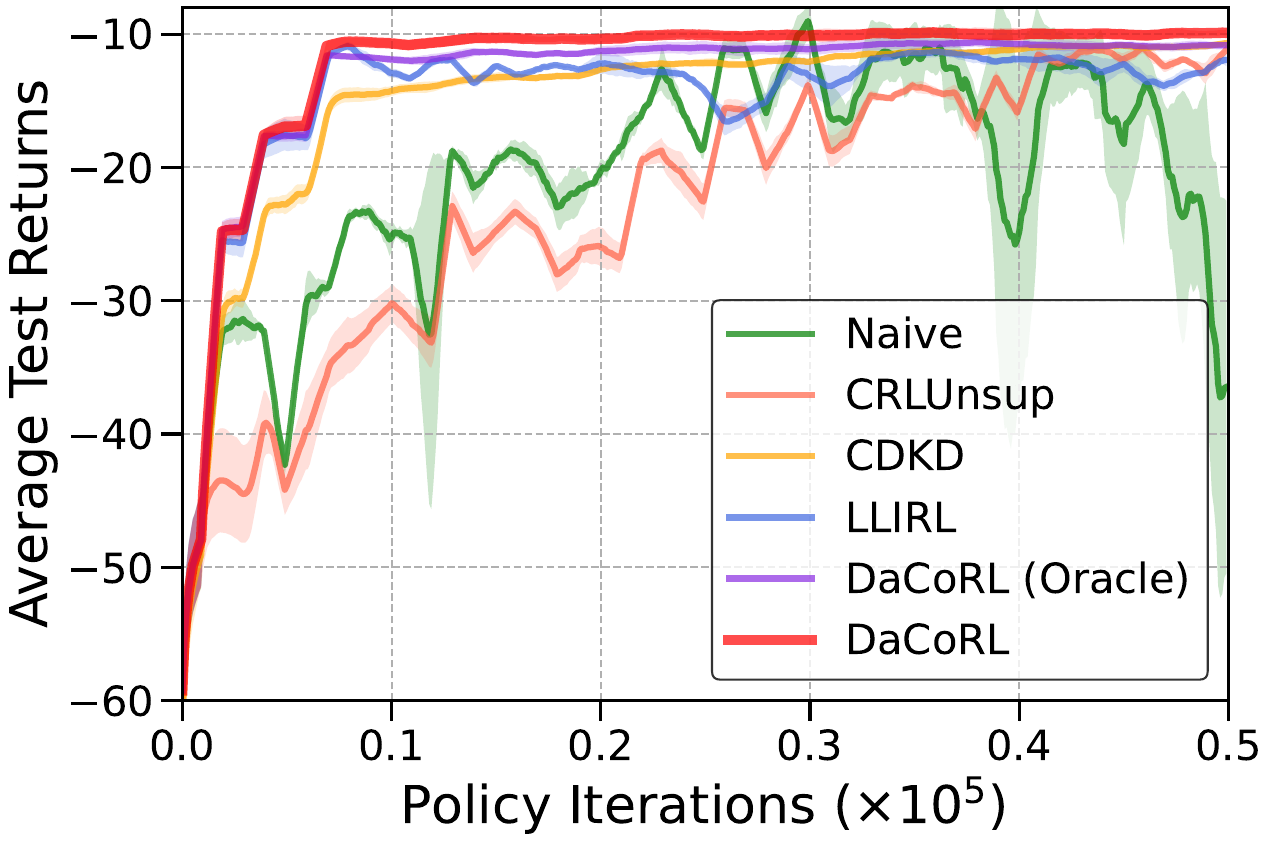}}
    \hspace{0.9cm}
    \subfigure[\label{fig:typeII}]
        {\includegraphics[width=0.27\linewidth]{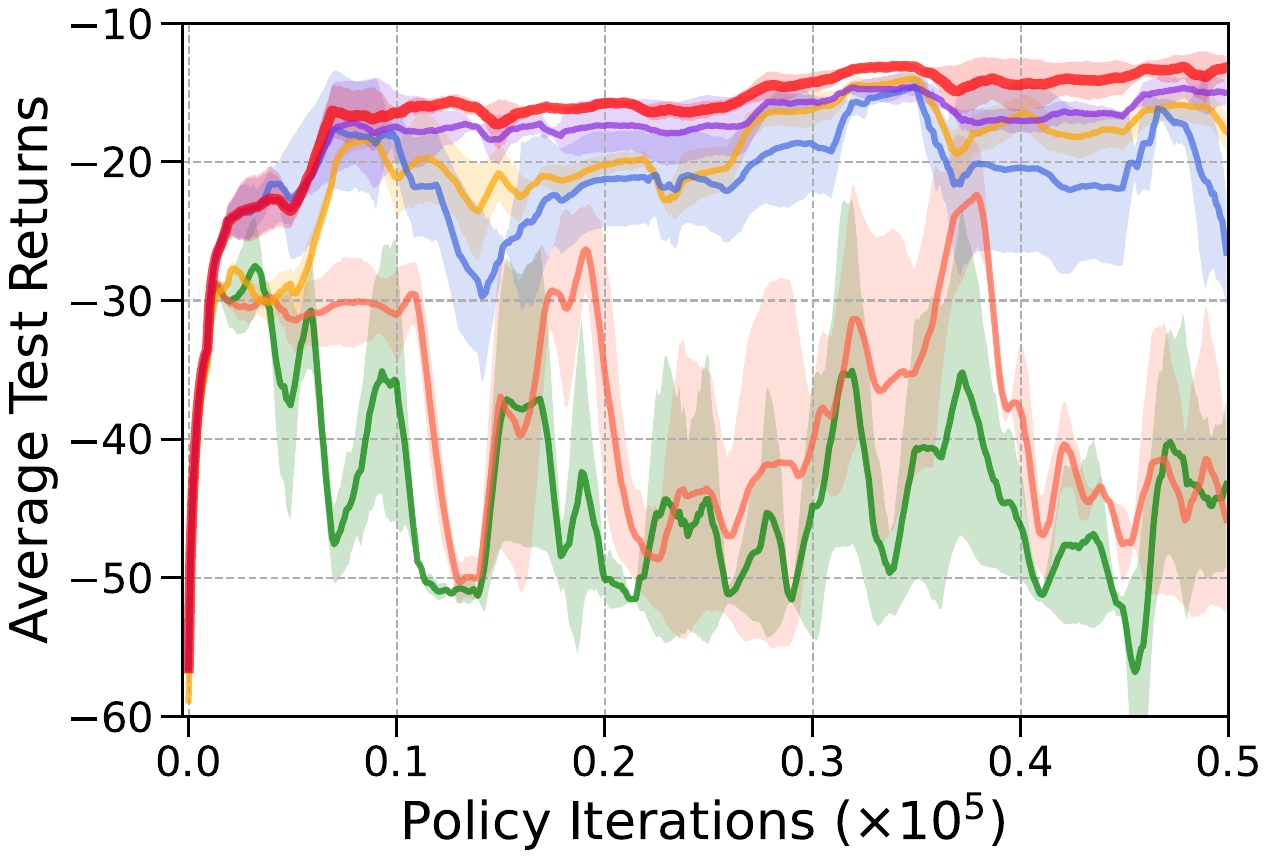}}
    \hspace{0.9cm}
    \subfigure[\label{fig:typeIII}]
        {\includegraphics[width=0.27\linewidth]{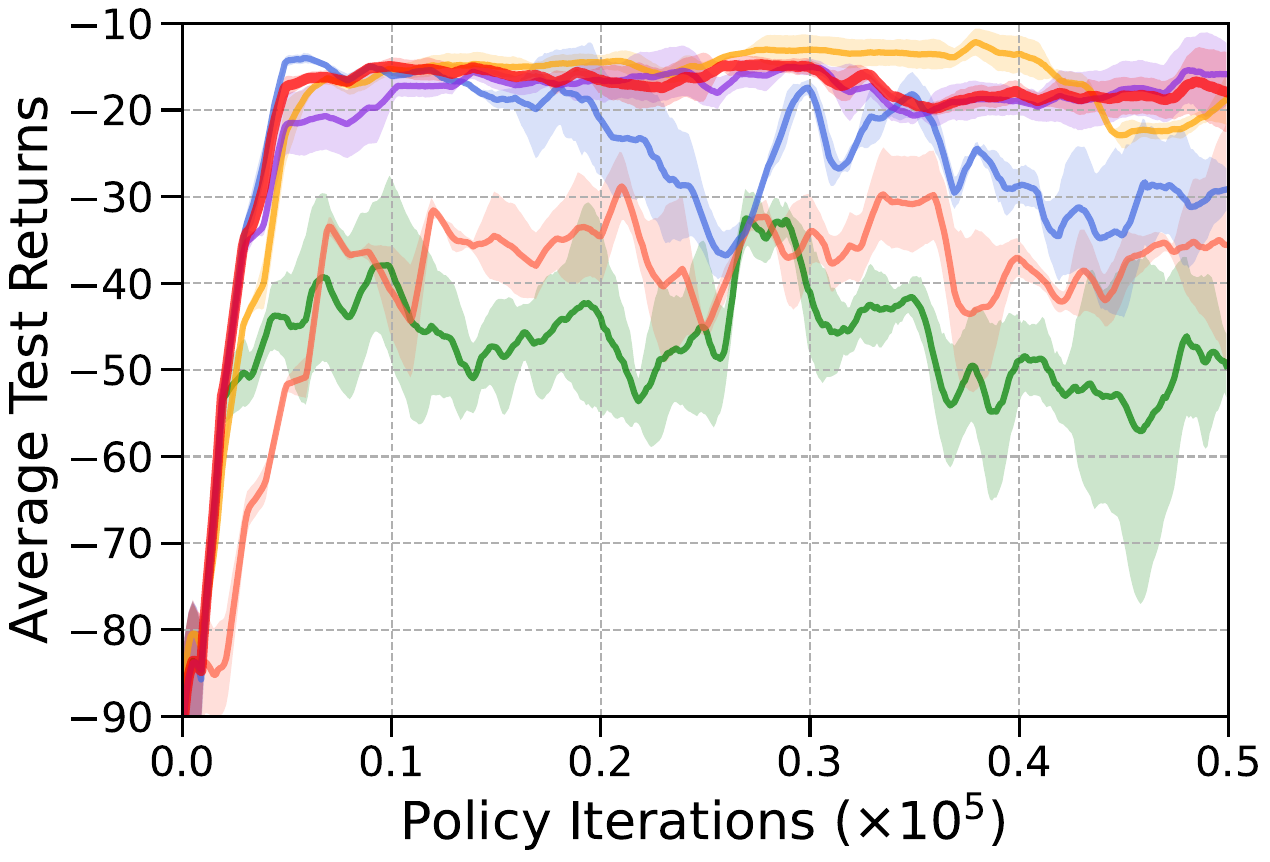}}
    \caption{Testing curves of the average returns over all tasks in the dynamic environments ($\mathcal{R}_{ave}$) of the navigation tasks. Here and in related figures in the following, $K_T$ is the number of instantiated contexts by DaCoRL and the solid lines and shaded regions denote the means and standard deviations of average test returns, respectively, across five runs. (a) Type I, $K_T=5$. (b) Type II, $K_T=6$. (c) Type III, $K_T=4$.}
    \label{fig:navi_performance}
\end{figure*}

\begin{figure}[t]
    \flushright
    \setlength{\abovecaptionskip}{3pt}
    \includegraphics[width=0.90\linewidth]{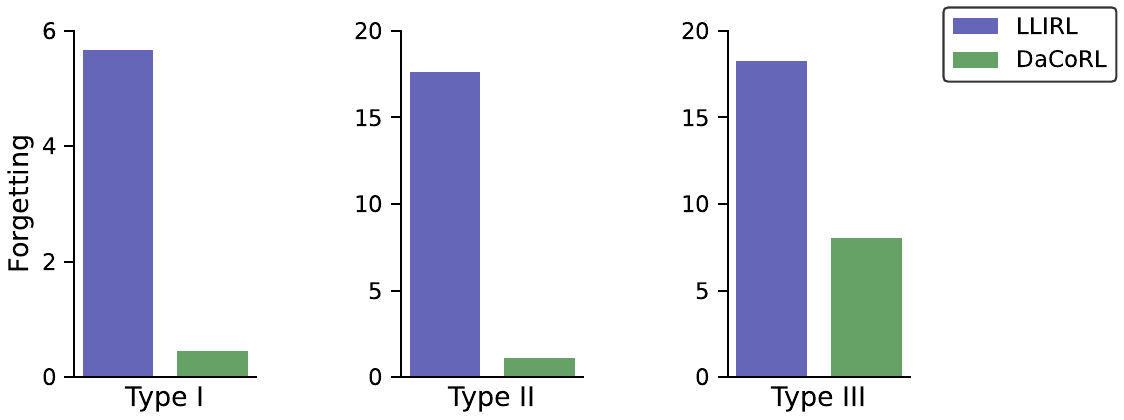}
    \caption{Average forgetting of LLIRL vs. DaCoRL in the navigation tasks. For each task $i$, we set forgetting as the decrease of performance after ending its training, $F_i=\mathcal{R}_i(i)-\mathcal{R}_i(T)$, where $\mathcal{R}_i(j)$ denotes the test returns for task $i$ after the $j^{\rm th}$ time period training. Here we report $F=\frac{1}{T}\sum_{i=1}^TF_i$. Positive value of $F$ indicates that the agent forgets a certain degree of learned information during the whole learning process, while the negative value indicates backward transfer.}
    \label{fig:navi_forgetting}
\end{figure}

\begin{table}[!t]
\centering
\setlength{\tabcolsep}{0.3mm}
\renewcommand\arraystretch{1.2}
\caption{Numerical results of $\bar{\mathcal{R}}_{ave}$ of all methods in the Navigation tasks (Based on the results in Fig. \ref{fig:navi_performance}. Here and in related\\ tables, the confidence intervals are standard devia-\\tions. The best performance is marked in boldface.)}
\label{table:navi_performance}
\begin{tabular}{c|c|c|c}
\toprule
Task               & Type I                  & Type II                        & Type III                  \\ \hline
Naive              & $-20.44\pm1.07$         & $-43.55\pm2.25$                & $-47.28\pm3.98$           \\
CRLUnsup           & $-22.43\pm0.97$         & $-37.73\pm1.82$                & $-40.32\pm1.02$           \\
CDKD               & $-14.34\pm0.30$         & $-19.86\pm0.69$                & \bm{$-19.13\pm0.95$}      \\
LLIRL              & $-14.24\pm0.24$         & $-21.08\pm1.92$                & $-25.92\pm0.41$           \\
DaCoRL             & \bm{$-11.93\pm0.43$}    &\bm{$-16.15\pm0.30$}            & $-19.58\pm0.08$           \\ \hline
DaCoRL (Oracle)     & $-12.79\pm0.26$         & $-17.61\pm0.03$                & $-20.46\pm1.19$           \\
\bottomrule
\end{tabular}
\end{table}


\subsection{Results}
To investigate the effectiveness of our proposed method (Q1), we present the results of DaCoRL and all baselines in three types of the navigation tasks. Fig. \ref{fig:navi_performance} shows the average episodic returns over all tasks during training according to \eqref{eq:R_test}, and Table \ref{table:navi_performance} reports the numerical results in terms of the average returns over $1000/100*50$ tests throughout the whole training process according to \eqref{eq:average_R_test}. 
For DaCoRL, the numbers of instantiated contexts are $K_T=5$, $K_T=6$, and $K_T=4$ for the three types of navigation tasks, respectively.
In Fig. \ref{fig:navi_performance}, it is clear that DaCoRL is significantly superior to all baselines in terms of the average test return and the stability. 
Naive shows the worst forgetting and performance fluctuations since it does not employ any mechanism to overcome data drifts in dynamic environments. 
CRLUnsup is slightly better than Naive due to the use of EWC technique. Its performance, however, is still far from that of other baselines due to the weak capacity to identify environmental changes just by looking at the cumulative reward curve.
CDKD and LLIRL perform clearly better than the above two baselines. Nevertheless, the fixed policy network structure may lead to an over-constrained optimization objective at the early stage of CDKD training, thus limiting the learning progress.
Although LLIRL employs separate neural networks to learn tasks from different contexts, which completely hinders the interference among between-context tasks, it involves more weights updating and model storage of neural networks and is primarily intended for faster adaptation as opposed to forgetting prevention. At the same time, it does not include any consideration of interference among tasks within the same context. 
As shown in Figure \ref{fig:navi_forgetting}, we visualize the the average performance degradation (called {\em forgetting}) of LLIRL and DaCoRL after ending each task's training and all sequential tasks' training in the navigation tasks. 
It is clear to see that compared with DaCoRL (train a multihead network using knowledge distillation technique), LLIRL training multiple separate context-specific neural networks has a substantially higher degree of forgetting of the learned task policies.

By contrast, due to the effective context division of dynamic environments in the latent space, along with an expandable multihead policy network and the knowledge distillation technique, DaCoRL can achieve competent continual learning of sequential tasks in dynamic environments.
In particular, it is on a par with [see Fig. \ref{fig:typeIII}] or even outperforms [see Figs. \ref{fig:typeI} and \ref{fig:typeII}] its supervised version [DaCoRL (Oracle)] where the task-to-context assignments are provided in advance. 
This results reveal that the incremental context detection module may produce more rational and valuable context division for continual learning than the predetermined task-to-context assignments.
Furthermore, the results in Table \ref{table:navi_performance} indicate that DaCoRL receives significantly larger or comparable average returns than all baselines during the entire learning process, and it also generally features smaller standard deviations in performance than the baselines.

\begin{figure*}[!t]
    \centering
    \setlength{\abovecaptionskip}{3pt}
    \subfigure[]
        {\includegraphics[width=0.27\linewidth]{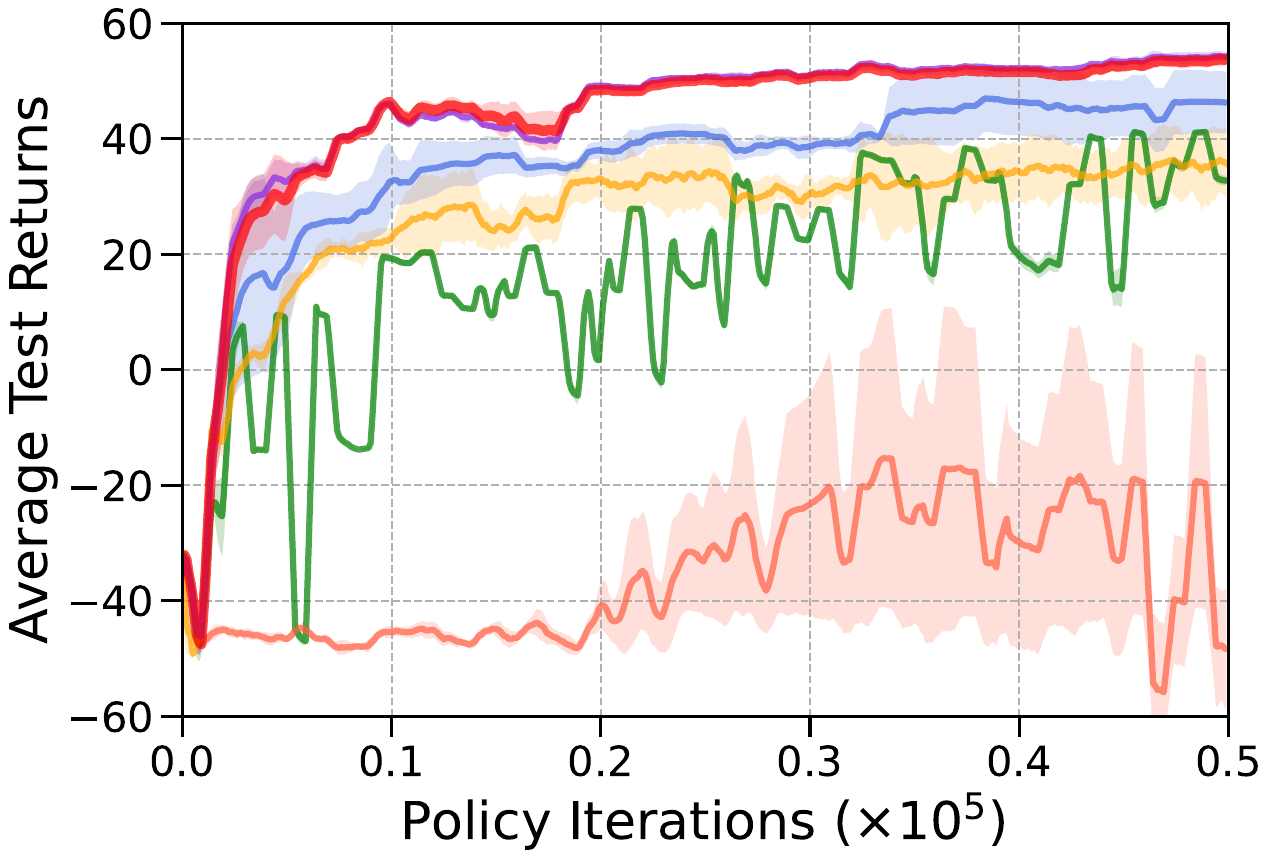}}
    \hspace{0.8cm}
    \subfigure[]
        {\includegraphics[width=0.27\linewidth]{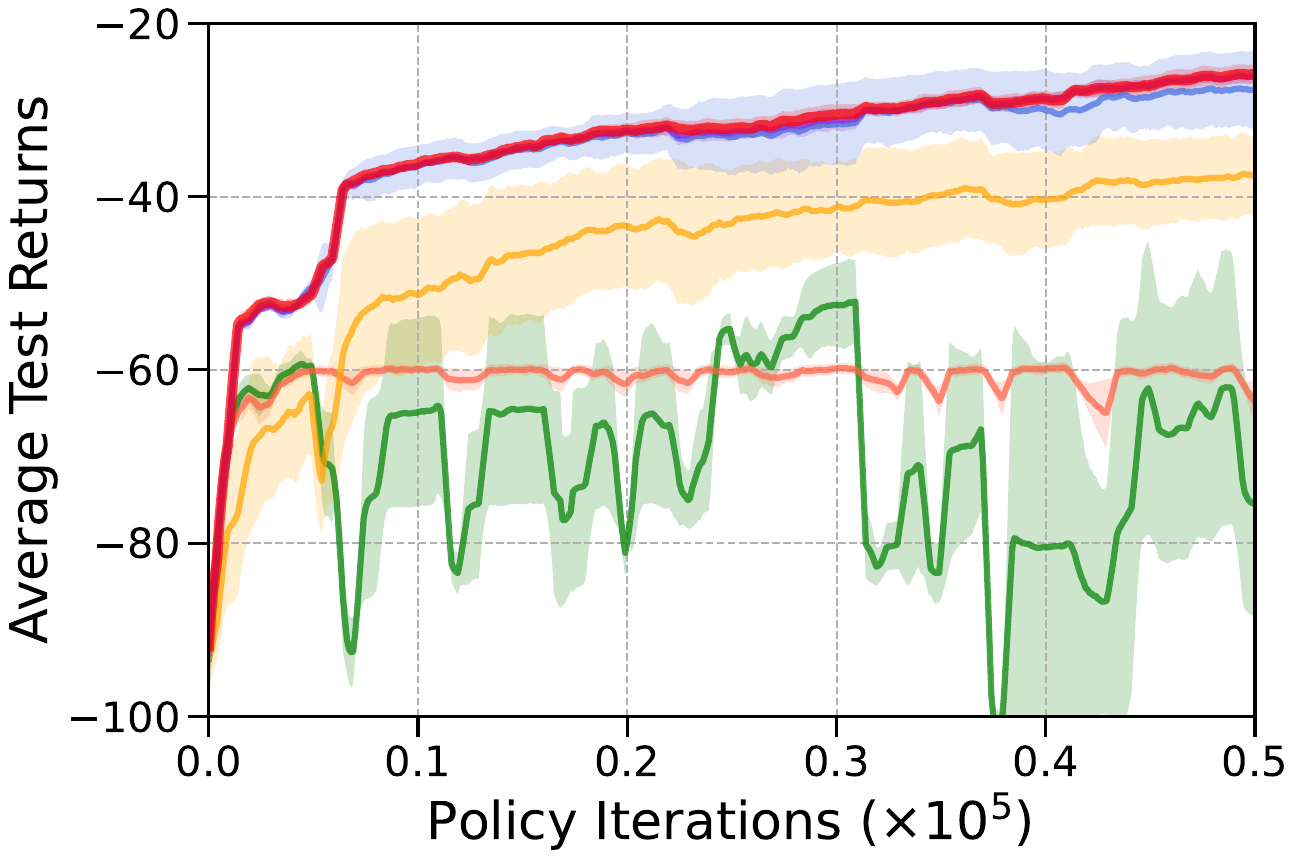}}
    \hspace{0.8cm}
    \subfigure[]
        {\includegraphics[width=0.27\linewidth]{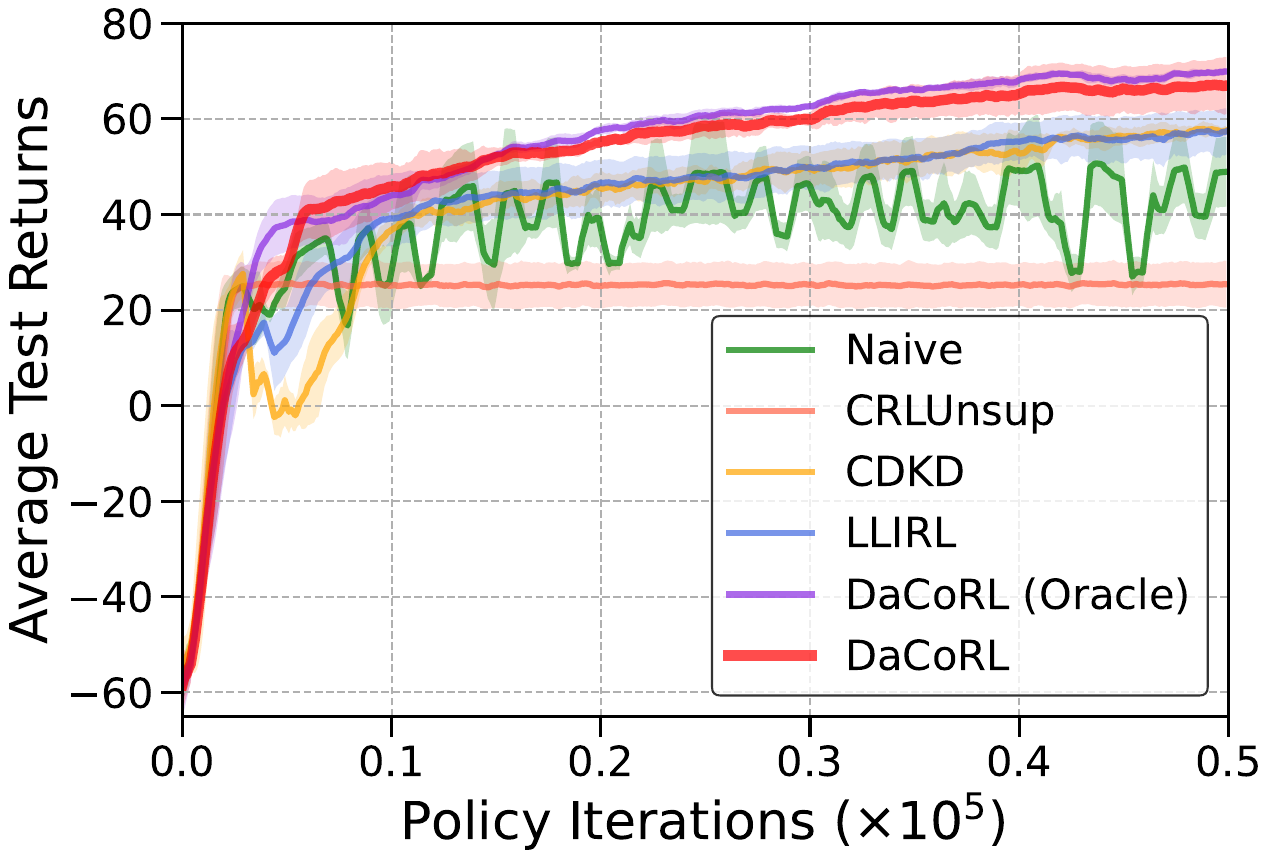}}
    \caption{Testing curves of the average returns over all tasks in the dynamic environments ($\mathcal{R}_{ave}$) of the MuJoCo locomotion tasks. $K_T=4$ contexts are instantiated for DaCoRL in all tasks. (a) Hopper. (b) HalfCheetah. (c) Ant.}
    \label{fig:mujoco_performance}
\end{figure*}

\begin{figure}[!t]
    \flushright
    \setlength{\abovecaptionskip}{3pt}
    \includegraphics[width=0.92\linewidth]{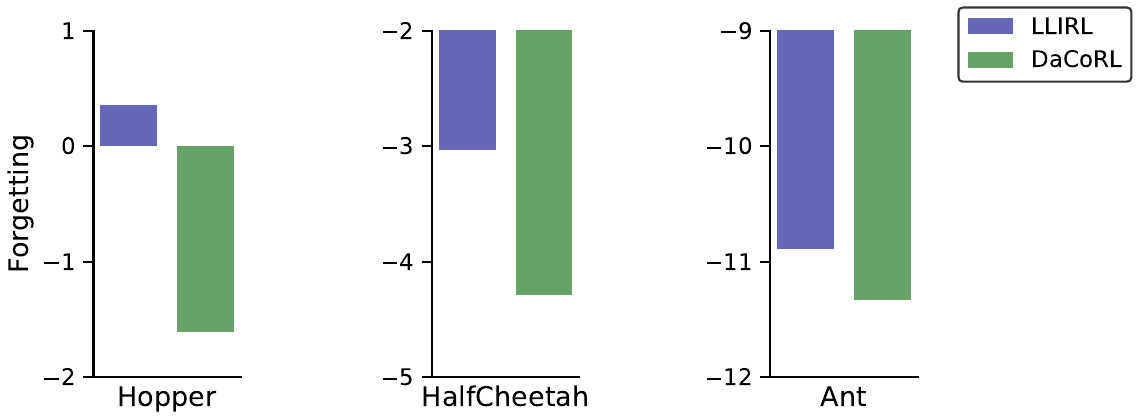}
    \caption{Statistics of average forgetting of LLIRL vs. DaCoRL in the MuJoCo locomotion tasks.}
    \label{fig:mujoco_forgetting}
\end{figure}

\begin{table}[!t]
\centering
\setlength{\tabcolsep}{0.8mm}
\renewcommand\arraystretch{1.2}
\caption{Numerical results of $\bar{\mathcal{R}}_{ave}$ of all methods in the MuJoCo Locomotion tasks (Based on the results in Fig. \ref{fig:mujoco_performance}.)}
\label{table:mujoco_performance}
\begin{tabular}{c|c|c|c}
\toprule
Task               & Hopper                     & HalfCheetah                   & Ant                       \\ \hline
Naive              & $16.18\pm0.32$             & $-69.98\pm6.44$               & $36.39\pm5.87$            \\
CRLUnsup           & $-35.87\pm9.38$            & $-61.08\pm0.55$               & $23.37\pm4.40$            \\
CDKD               & $26.47\pm3.73$             & $-46.36\pm6.63$               & $41.24\pm1.55$            \\
LLIRL              & $35.36\pm0.75$             & $-34.84\pm3.10$               & $42.76\pm4.25$            \\
DaCoRL             & \bm{$44.07\pm0.46$}        & \bm{$-34.11\pm0.62$}          & \bm{$52.18\pm1.34$}       \\ \hline
DaCoRL (Oracle)     & $44.57\pm0.45$             & $-34.31\pm0.58$                & $54.09\pm1.87$          \\
\bottomrule
\end{tabular}
\end{table}

To further demonstrate the scalability and flexibility of our method, we evaluate DaCoRL and all baselines on MuJoCo locomotion tasks.
All results are summarized in Fig. \ref{fig:mujoco_performance} and Table \ref{table:mujoco_performance}, from which we can see that DaCoRL consistently exhibits better and more stable performance than all baseline methods, even in these complex dynamic environments. More specifically, DaCoRL instantiates totally four contexts ($K_T=4$) for each dynamic environment, which is consistent with the given contexts in DaCoRL (Oracle). Since the learning curves of these two methods are largely coincident in all tasks in Fig. \ref{fig:mujoco_performance}, it further confirms that DaCoRL can accurately identify environmental context changes in a fully self-adaptive manner and can achieve comparable performance with the case where the task-to-context assignments are known in advance.
In addition, we also visualize the forgetting features of DaCoRL and LLIRL in this domain in Fig. \ref{fig:mujoco_forgetting}, which once again confirms the efficiency of DaCoRL in alleviating forgetting. 

\subsection{Analysis}
\label{analysis}

\begin{figure}[!t]
    \centering
    \setlength{\abovecaptionskip}{1pt}
    \subfigure[\label{TypeI_init}]
        {\includegraphics[width=0.3\linewidth]{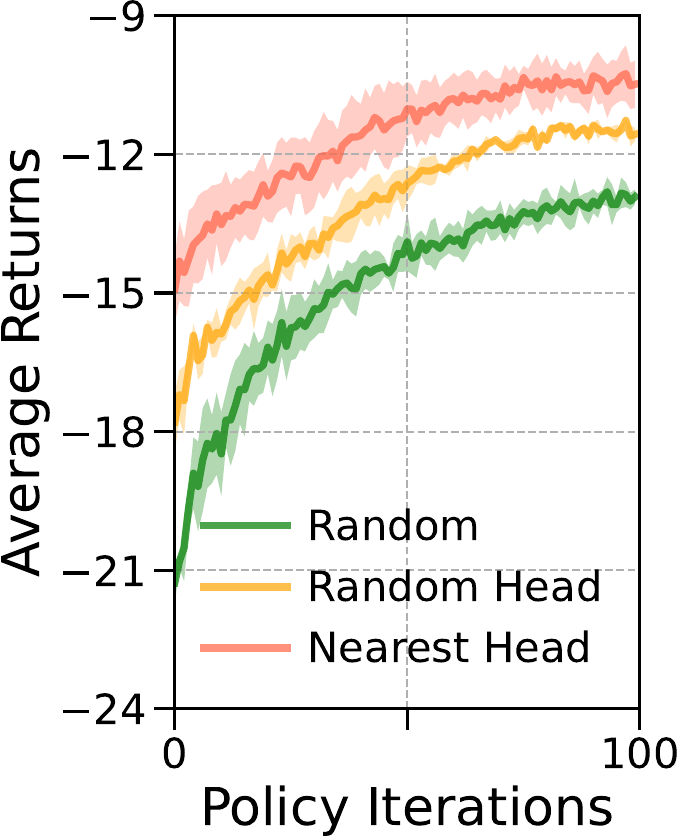}}
    \hspace{-0.1cm}
    \subfigure[\label{TypeII_init}]
        {\includegraphics[width=0.3\linewidth]{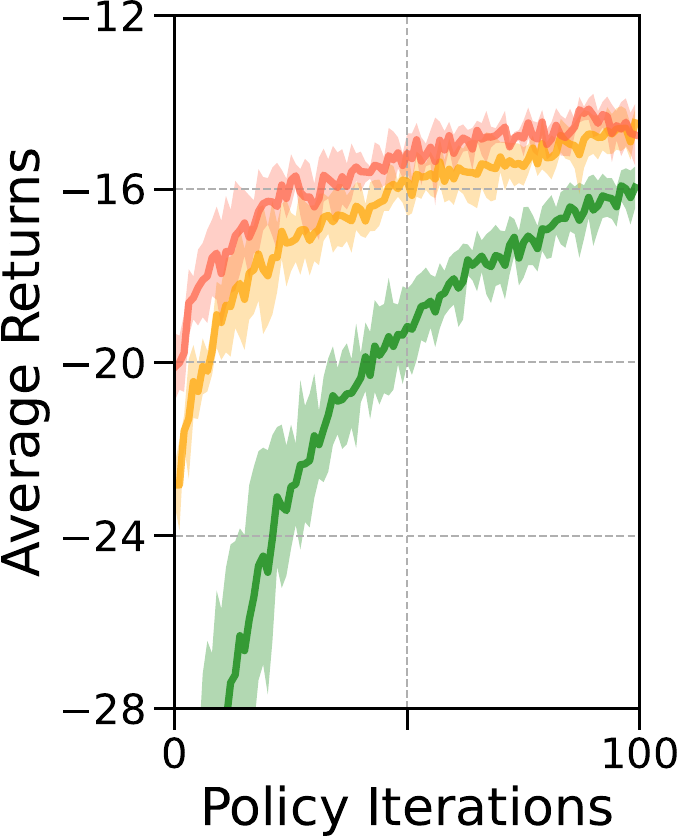}}
    \hspace{-0.1cm}
    \subfigure[\label{TypeIII_init}]
        {\includegraphics[width=0.3\linewidth]{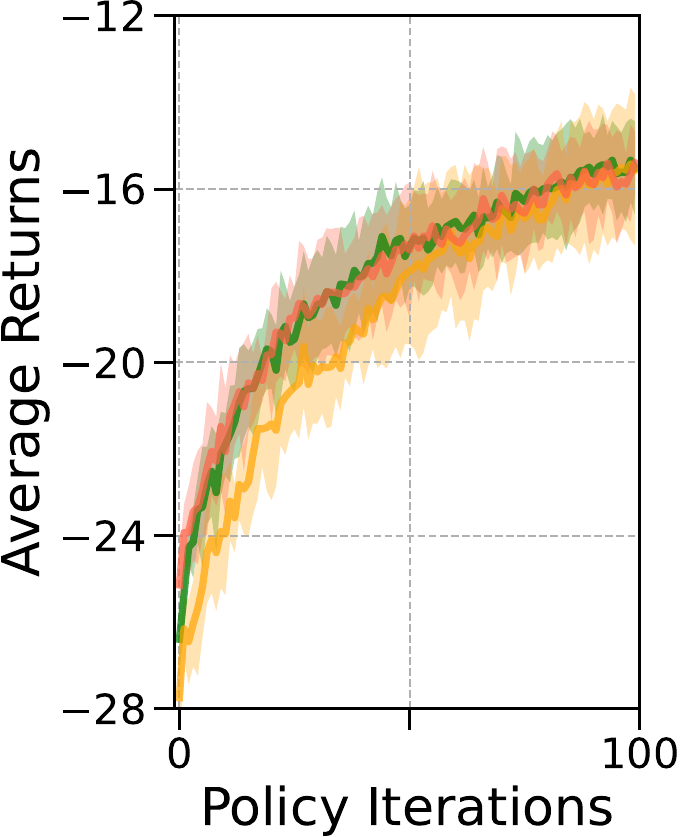}}
    \caption{Average training episodic return over all tasks per iteration of DaCoRL with different output head initialization implementations in the navigation tasks. (a) Type I. (b) Type II. (c) Type III.}
    \label{fig:performance_init}
\end{figure}

{\em 1) Influence of the Initialization of Output Heads}: 
To address Q2, we evaluate DaCoRL in the navigation tasks with different initialization strategies described in Section \ref{network_expansion}.
The average returns over all tasks during the first $100$ policy iterations are shown in Fig. \ref{fig:performance_init}. Here, we shorten the three initialization implementations to ``Random", ``Random Head" and ``Nearest Head", respectively, to ensure the readability.

From Fig. \ref{fig:performance_init}, it is clear that the initialization by Nearest Head can enable DaCoRL to attain a better initial policy and more positive forward transfer to the new task, which is consistent with our analysis in Section \ref{network_expansion}.  
Meanwhile, compared with Random initialization, Random Head exhibits obvious positive forward transfer on Type I and Type II environments and negative forward transfer on the Type III environment. Referring to \cite{wang2019incremental}, a possible explanation is that, in the Type III environment, Random Head often chooses an initial policy that hinders the exploration of the new task such that the agent needs to spend more time in the early training stage to counter the old policy, resulting in slow performance improvement.

\begin{figure*}[t]
    \centering
    \setlength{\abovecaptionskip}{3pt}
    \subfigure[]
        {\includegraphics[width=0.22\linewidth]{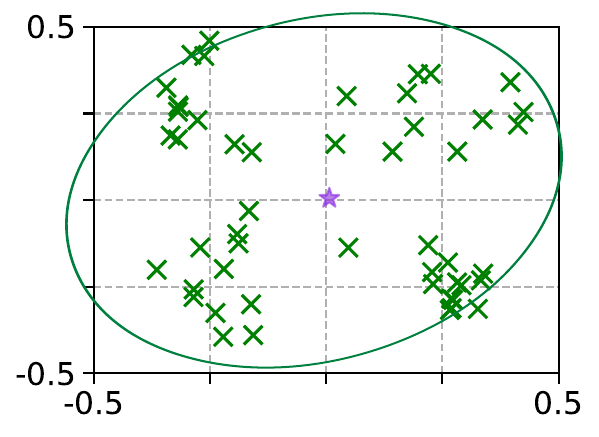}}
    \hspace{0.1cm}
    \subfigure[]
        {\includegraphics[width=0.22\linewidth]{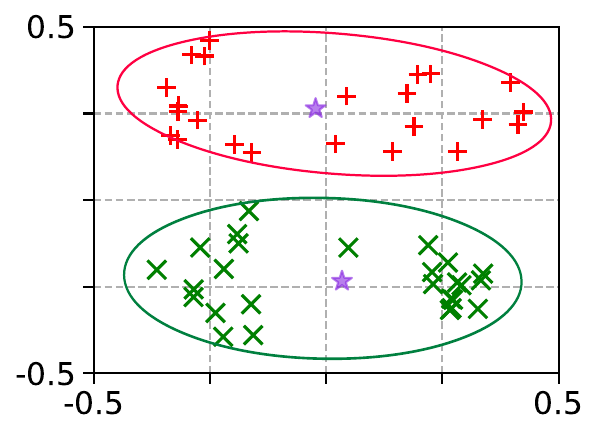}}
    \hspace{0.1cm}
    \subfigure[]
        {\includegraphics[width=0.22\linewidth]{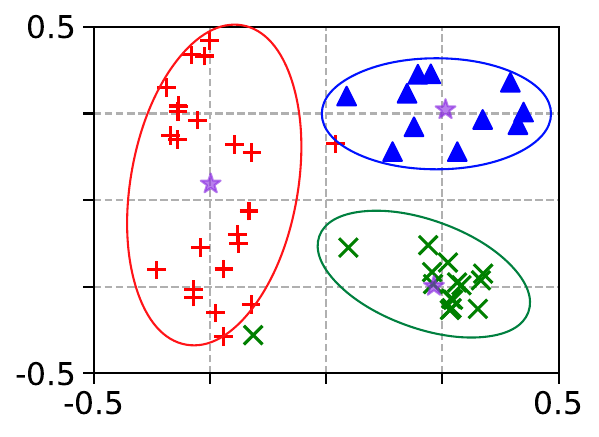}}
    \hspace{0.1cm}
    \subfigure[]
        {\includegraphics[width=0.22\linewidth]{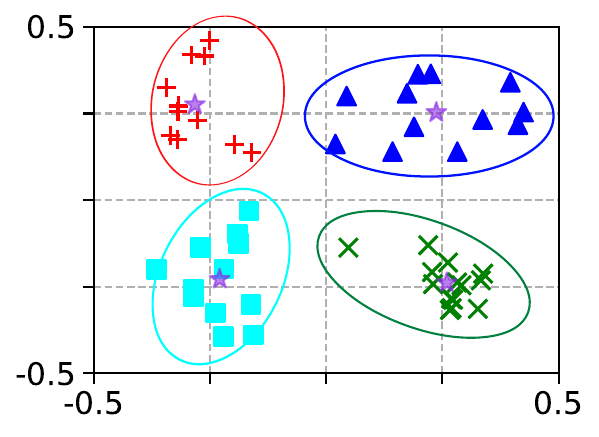}}
    \subfigure[]
        {\includegraphics[width=0.22\linewidth]{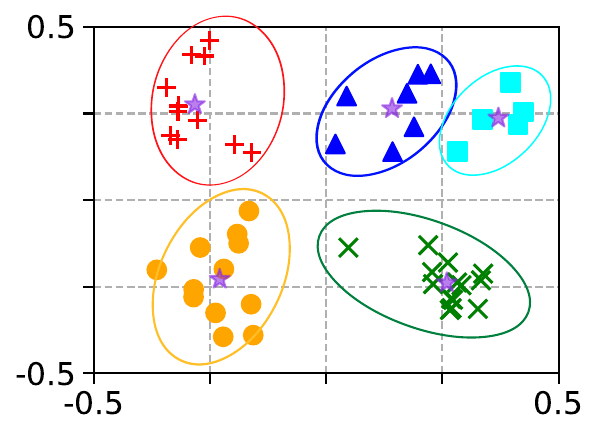}}
    \hspace{0.1cm}
    \subfigure[]
        {\includegraphics[width=0.22\linewidth]{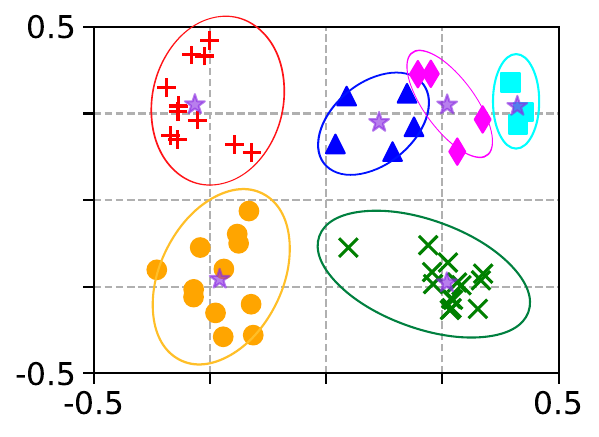}}
    \hspace{0.1cm}
    \subfigure[]
        {\includegraphics[width=0.22\linewidth]{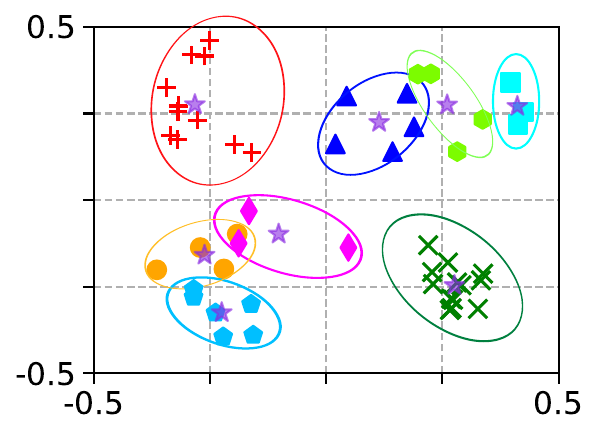}}
    \caption{The clustering patters of contexts in DaCoRL with different numbers of instantiated contexts in the Type I navigation environment. (a) One context. (b) Two contexts. (c) Three contexts. (d) Four contexts. (e) Five contexts. (f) Six contexts. (g) Eight contexts.}
    \label{fig:Instantiated Contexts}
\end{figure*}

\begin{figure*}[!t]
    \centering
    \setlength{\abovecaptionskip}{5pt}
    \subfigure[\label{TypeI_Contexts}]
        {\includegraphics[width=0.27\linewidth]{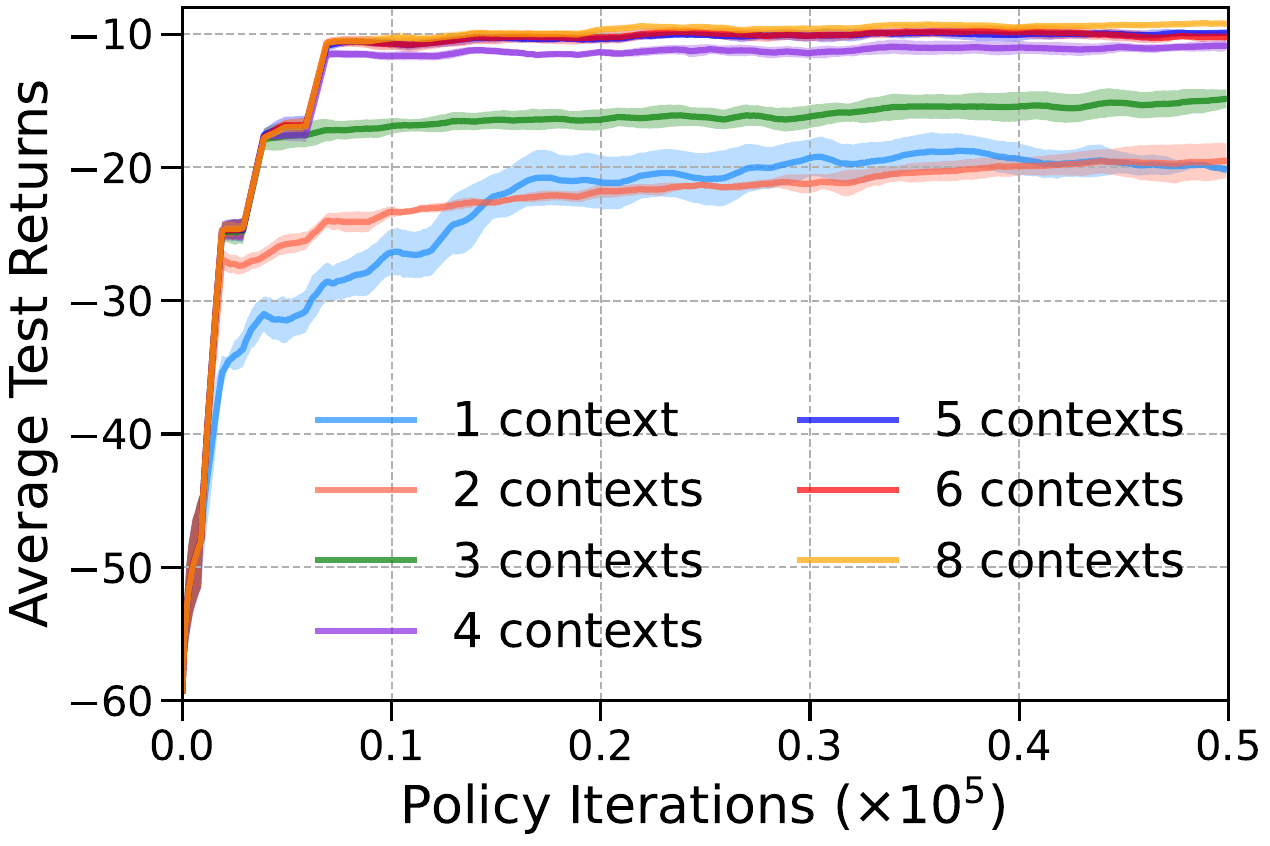}}
    \hspace{0.8cm}
    \subfigure[\label{TypeII_Contexts}]
        {\includegraphics[width=0.27\linewidth]{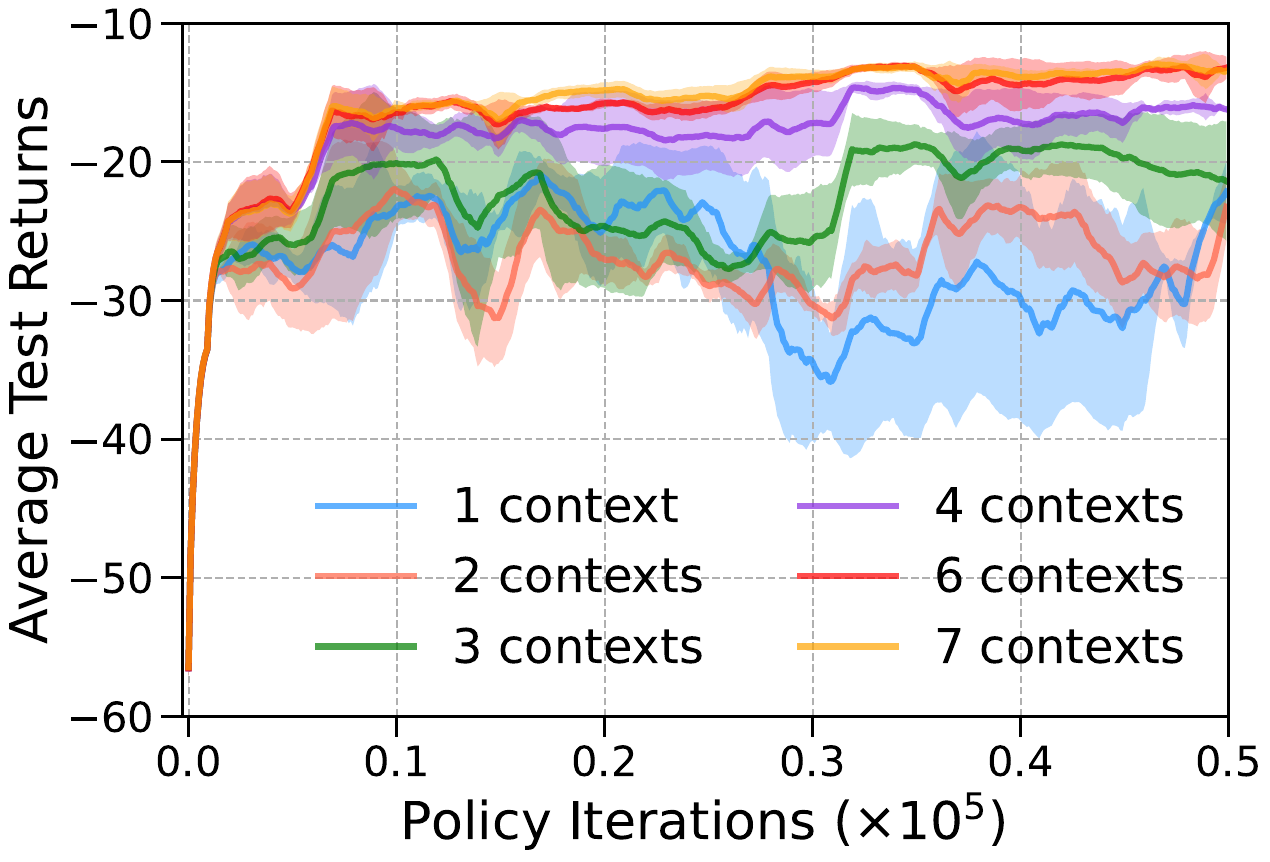}}
    \hspace{0.8cm}
    \subfigure[\label{TypeIII_Contexts}]
        {\includegraphics[width=0.27\linewidth]{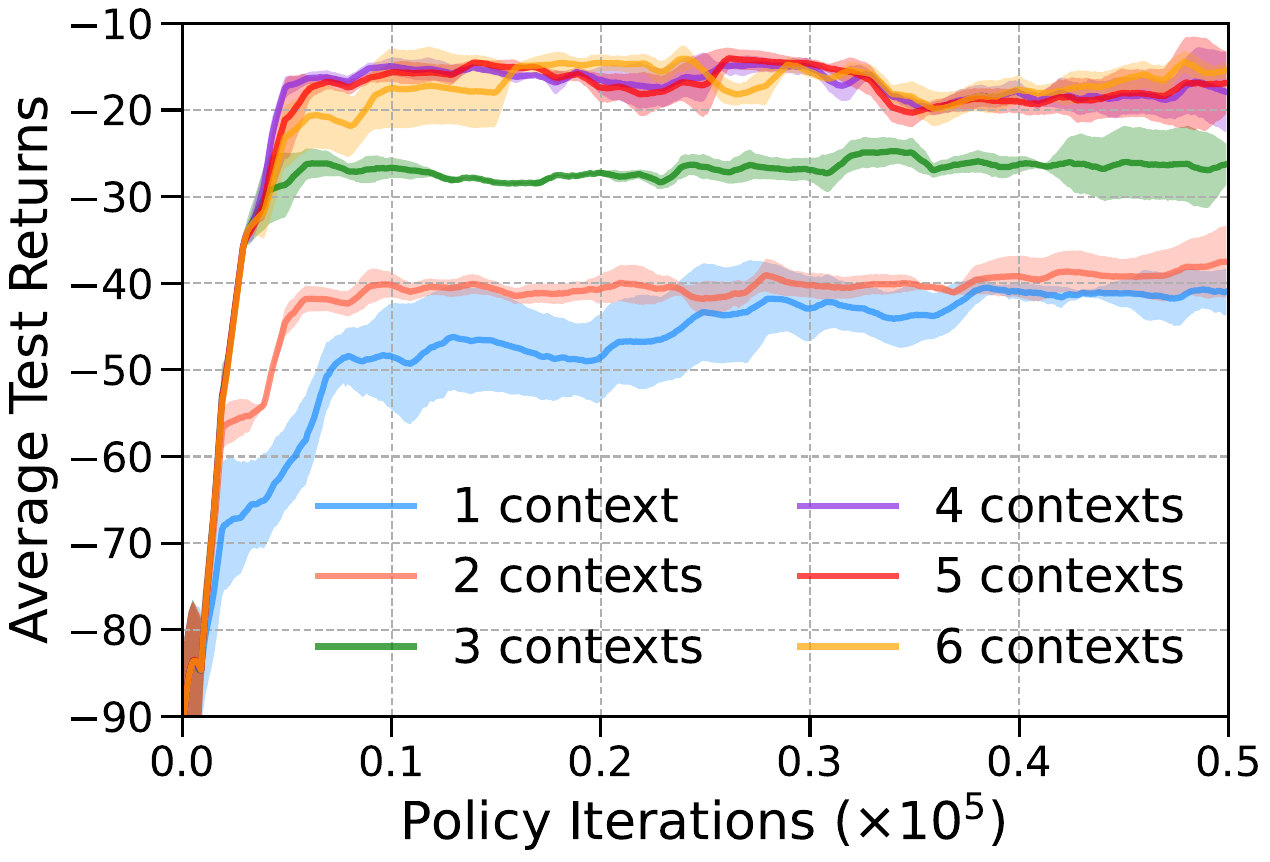}}
    \caption{The average returns over all tasks in the dynamic environments ($\mathcal{R}_{ave}$) of DaCoRL with different numbers of instantiated contexts in the navigation tasks. (a) Type I. (b) Type II. (c) Type III.}
    \label{fig:performance_Instantiated_Contexts}
\end{figure*}

{\em 2) Influence of the Number of Instantiated Contexts ($K_T$)}: 
Intuitively, instantiating a separate context for each task contained in the dynamic environment (i.e., $K_T=T$, task-specific output head within an individual time period) is likely to result in an optimal policy for the agent. 
However, because of the intricate network architecture and challenging model training procedure, it is problematic to use in practice, particularly when the dynamic environment comprises a large number of tasks 
Thus, it is necessary to investigate the influence of the number of contexts on the performance of DaCoRL (Q3).

We change the numbers of instantiated contexts in DaCoRL by varying the concentration parameter $\alpha$ in the CRP in the navigation tasks. 
Since the environmental features are highly correlated with their variation parameters, that is, intuitively, taking the Type I environment for example, the tasks with adjacent goal positions are more similar to each other and tend to belong to the same context. Hence, we use the goal position in the 2-D coordinate as a visualization to reveal the clustering patterns of contexts for the Type I environment.
The results are shown in Fig. \ref{fig:Instantiated Contexts}, where the star shapes represent the centroids of instantiated contexts obtained from the incremental context detection procedure, and other markers represent tasks ($M_i$, $i\in[1,2,\dots,T]$) in the dynamic environment. It is clear that our incremental context detection module can capture the appropriate context patterns under different concentration parameter settings in a fully online manner, regardless of how many contexts are ultimately instantiated.

\begin{table}[!t]
\centering
\setlength{\tabcolsep}{1.8mm}
\renewcommand\arraystretch{1.2}
\caption{Numerical results of $\bar{\mathcal{R}}_{ave}$ of DaCoRL with different\\ numbers of instantiated contexts in the robot\\ Navigation tasks (Based on the results in Fig. \ref{fig:performance_Instantiated_Contexts}.)}
\label{table:performance_Instantiated_Contexts}
\begin{tabular}{c|c|c|c}
\toprule
$K_T$             & Type I                   & Type II                  & Type III               \\ \hline
1                 & $-22.86\pm1.10$          & $-27.34\pm3.42$          & $-47.16\pm3.69$        \\                                   
2                 & $-22.32\pm0.34$          & $-26.80\pm0.43$          & $-42.08\pm1.35$         \\
3                 & $-17.09\pm0.88$          & $-22.59\pm2.10$          & $-28.74\pm0.89$        \\
4                 & $-12.89\pm0.30$          & $-18.00\pm0.96$          & \bm{$-19.58\pm0.08$}   \\
5                 & \bm{$-11.93\pm0.43$}     & $-$                      & \bm{$-19.80\pm0.42$}   \\
6                 & \bm{$-11.89\pm0.43$}     & \bm{$-16.15\pm0.30$}     & \bm{$-19.86\pm0.56$}   \\
7                 & $-$                      & \bm{$-15.79\pm0.21$}     & $-$                    \\
8                 & \bm{$-11.48\pm0.19$}     & $-$                      & $-$                    \\
\bottomrule
\end{tabular}
\end{table}

Additionally, we show the performance of DaCoRL with different numbers of instantiated contexts in the three types of navigation environments in Fig. \ref{fig:performance_Instantiated_Contexts} and Table \ref{table:performance_Instantiated_Contexts} according to \eqref{eq:R_test} and \eqref{eq:average_R_test}, respectively. 
Overall, the results are consistent with the intuition that DaCoRL tends to achieve better performance with more contexts instantiated. Meanwhile, maintaining fewer contexts means that tasks with large differences may be clustered into the same context, leading to more interference during policy network training. 
Nevertheless, it should be noted that, in Fig. \ref{fig:performance_Instantiated_Contexts} and Table \ref{table:performance_Instantiated_Contexts}, the performance gains become relatively minor when the number of contexts exceeds a certain value. Consequently, for the sake of model complexity and training cost, we recommend choosing a moderate number of instantiated contexts by adjusting the value of $\alpha$. For instance, in our experiments, $K_T=5$ for Type I, $K_T=6$ for Type II and $K_T=4$ for Type III are sufficient to obtain reasonable performance in these navigation tasks. 

We also record the rough value range of the concentration parameter $\alpha$ related to the specific number of contexts. The results are presented in Appendix \ref{appdix:concentration_paras} from which we empirically found that this hyperparameter is relatively insensitive, and all $\alpha$ values within the corresponding continuous interval can result in the same context instantiation results. This insensitivity ensures a certain degree of simplicity and convenience of parameter tuning in practical implementations.

\begin{table*}[t]
\centering
\setlength{\tabcolsep}{3.mm}
\renewcommand\arraystretch{1.19}
\caption{Numerical results in terms of the average initial performance over all sequential tasks\\ during training in the robot Navigation and MuJoCo Locomotion tasks}
\label{table:Forward_Transfer}
\begin{tabular}{c|c|c|c|c|c|c}
\toprule
Task       & Type I                 & Type II                & Type III               & Hopper                & HalfCheetah            & Ant                   \\ \hline
Naive      & $-22.17\pm1.83$        & $-43.63\pm2.42$        & $-50.86\pm5.18$        & $8.67\pm0.47$         & $-75.11\pm4.57$        & $30.64\pm4.39$        \\
CRLUnsup   & $-23.07\pm1.23$        & $-39.09\pm0.58$        & $-44.78\pm0.90$        & $-39.40\pm9.42$       & $-63.55\pm0.47$        & $20.91\pm4.56$        \\
CDKD       & $-20.11\pm0.60$        & $-31.34\pm1.77$        & $-25.96\pm0.86$        & $24.42\pm3.93$        & $-49.66\pm7.64$        & $34.57\pm2.43$        \\
LLIRL      & $-19.08\pm0.51$        & $-23.80\pm1.71$        & $-33.47\pm1.36$        & $28.16\pm0.77$        & $-37.63\pm3.55$        & $40.53\pm3.70$        \\
DaCoRL     & \bm{$-14.92\pm0.74$}   & \bm{$-20.11\pm0.78$}   & \bm{$-25.14\pm0.67$}   & \bm{$37.28\pm0.46$}   & \bm{$-36.02\pm0.13$}   & \bm{$49.07\pm1.38$}   \\
\bottomrule
\end{tabular}
\end{table*}

\begin{table*}[!t]
\centering
\setlength{\tabcolsep}{3.mm}
\renewcommand\arraystretch{1.19}
\caption{Numerical results in terms of the average test return over 50 different tasks that have not\\ been seen during training in robot Navigation and MuJoCo Locomotion tasks}
\label{table:Generalization}
\begin{tabular}{c|c|c|c|c|c|c}
\toprule
Task       & Type I                 & Type II               & Type III               & Hopper                & HalfCheetah           & Ant                     \\ \hline
Naive      & $-33.78\pm19.51$       & $-39.82\pm9.78$       & $-50.19\pm8.02$        & $39.36\pm0.90$        & $-69.84\pm12.17$      & $55.26\pm6.01$          \\
CRLUnsup   & $-11.29\pm0.59$        & $-42.52\pm9.09$       & $-38.24\pm16.41$       & $-43.41\pm12.73$      & $-62.92\pm1.57$       & $34.13\pm4.34$          \\
CDKD       & $-11.86\pm0.24$        & $-22.46\pm2.06$       & \bm{$-18.77\pm3.93$}   & $34.25\pm9.38$        & $-36.14\pm4.05$       & $63.26\pm1.86$          \\
LLIRL      & $-14.52\pm1.28$        & $-39.84\pm1.26$       & $-35.94\pm0.15$        & $48.76\pm3.39$        & $-27.67\pm3.85$       & $59.39\pm2.25$          \\
DaCoRL     & \bm{$-11.28\pm0.22$}   & \bm{$-17.58\pm0.89$}  & $-23.38\pm4.46$        & \bm{$53.45\pm0.90$}   & \bm{$-26.36\pm1.20$}  & \bm{$67.65\pm4.24$}     \\
\bottomrule
\end{tabular}
\end{table*}

{\em 3) Forward Transfer}:
In DaCoRL, tasks with similar dynamics are grouped into the same context and share the same policy throughout the CRL process. In each time period, DaCoRL retrieves the policy corresponding to the most similar context as the initial policy for the current learning process, which can enable the agent to get better startup performance on each new task.
To validate this feature (Q4), we enumerate the average training episodic return for each algorithm at the first policy iteration on each task, and the results are shown in Table \ref{table:Forward_Transfer}.
In comparison to all baselines, DaCoRL consistently achieves better initial startup performance in all experimental dynamic environments, displaying a strong positive forward transfer ability.

{\em 4) Generalization Results}: 
To investigate the generalization performance of DaCoRL in unseen tasks (Q5), we randomly generate $50$ different tasks that follow the same distribution as the training tasks but have not been seen during training in each dynamic environment. Then, we test the policies learned by DaCoRL and all baselines in these tasks, where each policy is tested on a specific task for $100$ times. The average test returns over all tasks are shown in Table \ref{table:Generalization}.
In general, DaCoRL features superior generalization ability in most unseen tasks compared with all baselines, except on Type III navigation tasks, where its generalization performance is slightly inferior to CDKD but still significantly better than other three baselines. 
This phenomenon is explainable since CDKD is the most similar algorithm to DaCoRL and is likely to be comparable to DaCoRL especially when the numbers of contexts in both methods are consistent.

{\em 5) Sample Efficiency}: Compared with Naive (no consideration of environmental changes) and CRLUnsup (identifying environmental changes by evaluating the cumulative rewards during training), methods with a separate context detection phase (i.e., CDKD, LLIRL and DaCoRL) are required to explore the current environment using a uniform policy and collect extra samples for context inference. 
We record the number of agent-environment interaction episodes during training on a sequential tasks for all approaches, and the results are summarized in Appendix \ref{appdix:sample_efficiency}. 
Combined with the experimental results shown in the preceding part of this section, it can be concluded that: 
i) Compared with Naive and CRLUnsup, DaCoRL sacrifices a certain degree of sample efficiency, but achieves great performance improvement; 
ii) The performance of DaCoRL is also significantly better than that of CDKD and LLIRL, although the sample efficiency of these three methods is consistent.

\section{Conclusion and Future Work}
In this article, we present a CRL framework named DaCoRL as a viable solution for the agent to successfully conduct continual learning in dynamic environments.
The goal of DaCoRL is to continuously adapt the RL agent's behavior towards the changing environment, and to minimize the catastrophic forgetting of previously learned tasks.
To this end, we employs an incremental context detection module to categorize the stream of tasks into a set of distinct contexts using online Bayesian infinite Gaussian mixture clustering. Then, a context-conditioned policy with an expandable multihead neural network are optimized, in conjunction with a knowledge distillation regularization term, to avoid the interference among tasks both between and within contexts.
A key profit of our method is that it can achieve incremental context instantiation in a fully online manner without requiring any external information to explicitly signal environmental changes. Meanwhile, it only relies on a single policy network to accomplish effective continual learning in dynamic environments and can be easily coupled with any other policy-based RL algorithms. Experiments on several continuous control tasks confirm that DaCoRL can significantly outperform state-of-the-art algorithms in terms of the stability, overall performance and generalization ability.

This article mainly address the dynamic environment scenarios with abrupt changes between tasks. A promising direction for future work is to explore more challenging cases, including subtle variation from one task to another, or intensively changing environments where the changes may happen between consecutive episodes.
Another direction is to develop an efficient strategy that can facilitate the forward and backward transfer during the CRL training process.

\bibliographystyle{IEEEtran}
\bibliography{IEEEabrv,IEEEexample}


\vspace{-1.1cm}
\begin{IEEEbiography}[{\includegraphics[width=1in,height=1.25in,clip,keepaspectratio]{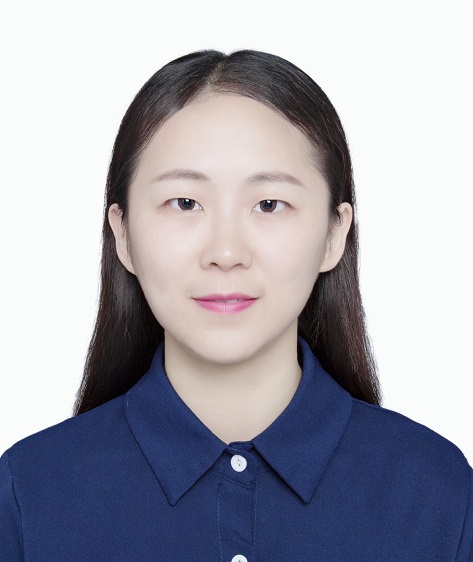}}]{Tiantian Zhang}
received the B.Sc. degree in automation from the Department of Information Science and Technology, Central South University, Changsha, China, in 2015, and the M.Sc. degree in control engineering from the Department of Automation, Tsinghua University, Beijing, China, in 2018, where she is currently pursuing the Ph.D. degree in control science and engineering. 

Her research interests include data science, decision-making, and reinforcement learning.
\end{IEEEbiography}

\vspace{-1.1cm}
\begin{IEEEbiography}[{\includegraphics[width=1in,height=1.25in,clip,keepaspectratio]{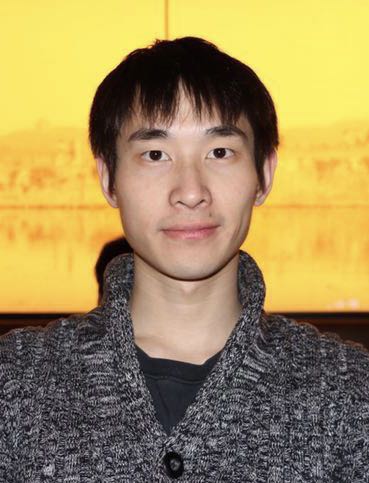}}]{Zichuan Lin}
finished his Ph.D. degree from the department of Computer Science and Technology, Tsinghua University, Beijing, China, in 2021. 

He is now a Researcher at Tencent, Shenzhen, China, working on sample-efficient deep reinforcement learning algorithms. He is interested in machine learning, reinforcement learning as well as their applications on game AI and natural language processing. He has been serving as a PC for NeurIPS, ICML, ICLR and AAAI. 
\end{IEEEbiography}

\vspace{-1.1cm}
\begin{IEEEbiography}[{\includegraphics[width=1in,height=1.25in,clip,keepaspectratio]{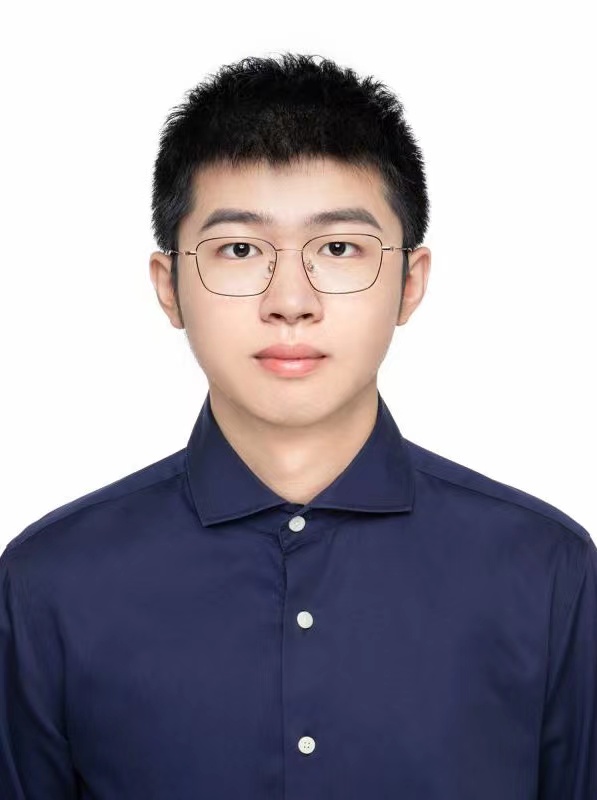}}]{Yuxing Wang} received the B.Sc. degree in communication engineering from the Department of Electronic Information, Southwest Minzu University, Chengdu, China, in 2020. He is currently pursuing the M.Sc. degree in electronic information with the Shenzhen International Graduate School, Tsinghua University, Shenzhen, China. 

His research interests include evolutionary computation, reinforcement learning and embodied AI.
\end{IEEEbiography}

\vspace{-1.1cm}
\begin{IEEEbiography}[{\includegraphics[width=1in,height=1.25in,clip,keepaspectratio]{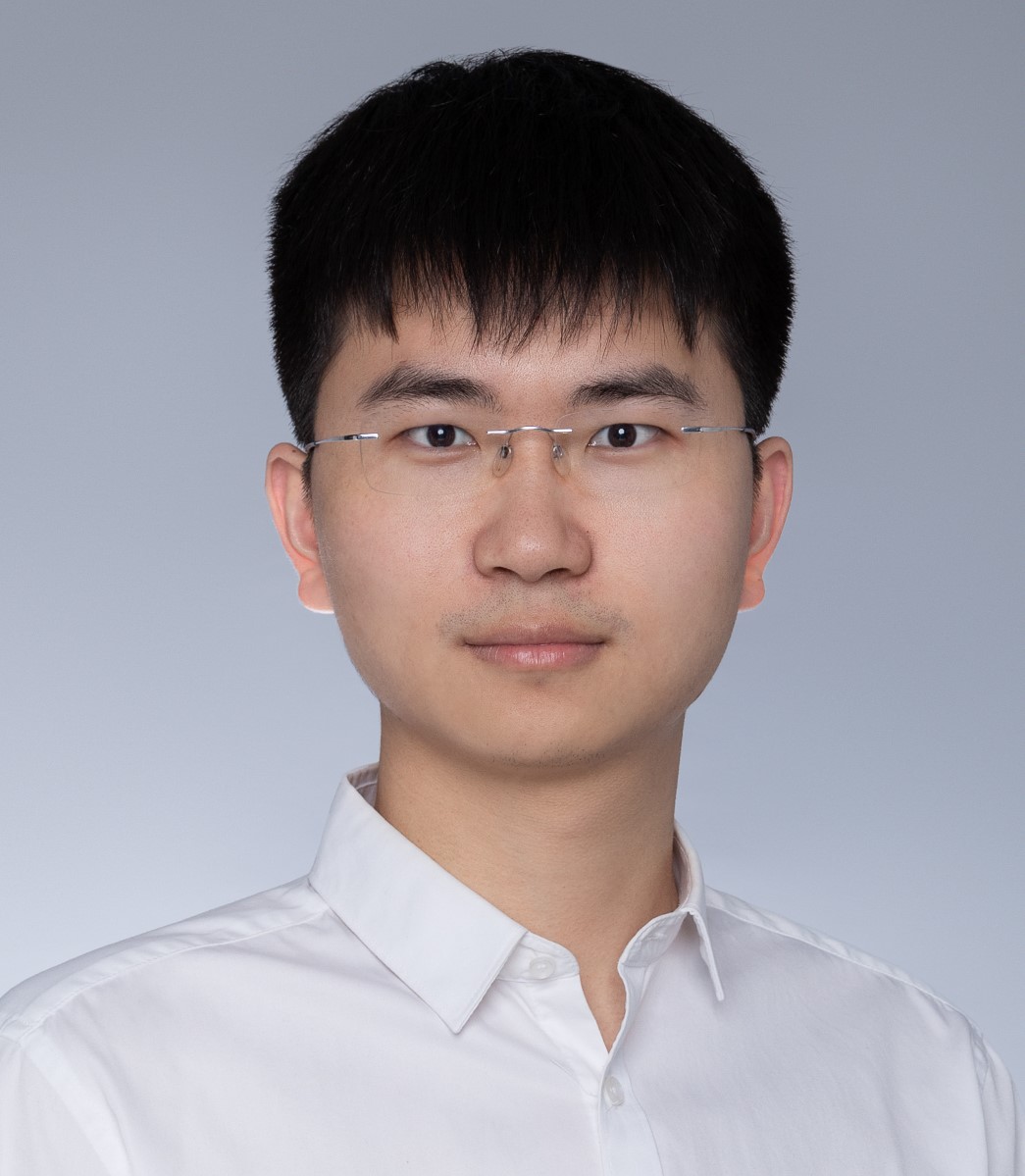}}]{Deheng Ye} finished his Ph.D. from the School of Computer Science and Engineering, Nanyang Technological University, Singapore, in 2016. 

He is now a Principal Researcher and Team Manager with Tencent, Shenzhen, China, where he leads a group of engineers and researchers developing large-scale learning platforms and intelligent AI agents. He is interested in applied machine learning, reinforcement learning, and software engineering. He has been serving as a PC/SPC for NeurIPS, ICML, ICLR, AAAI, and IJCAI.
\end{IEEEbiography}

\vspace{-1.1cm}
\begin{IEEEbiography}[{\includegraphics[width=1in,height=1.25in,clip,keepaspectratio]{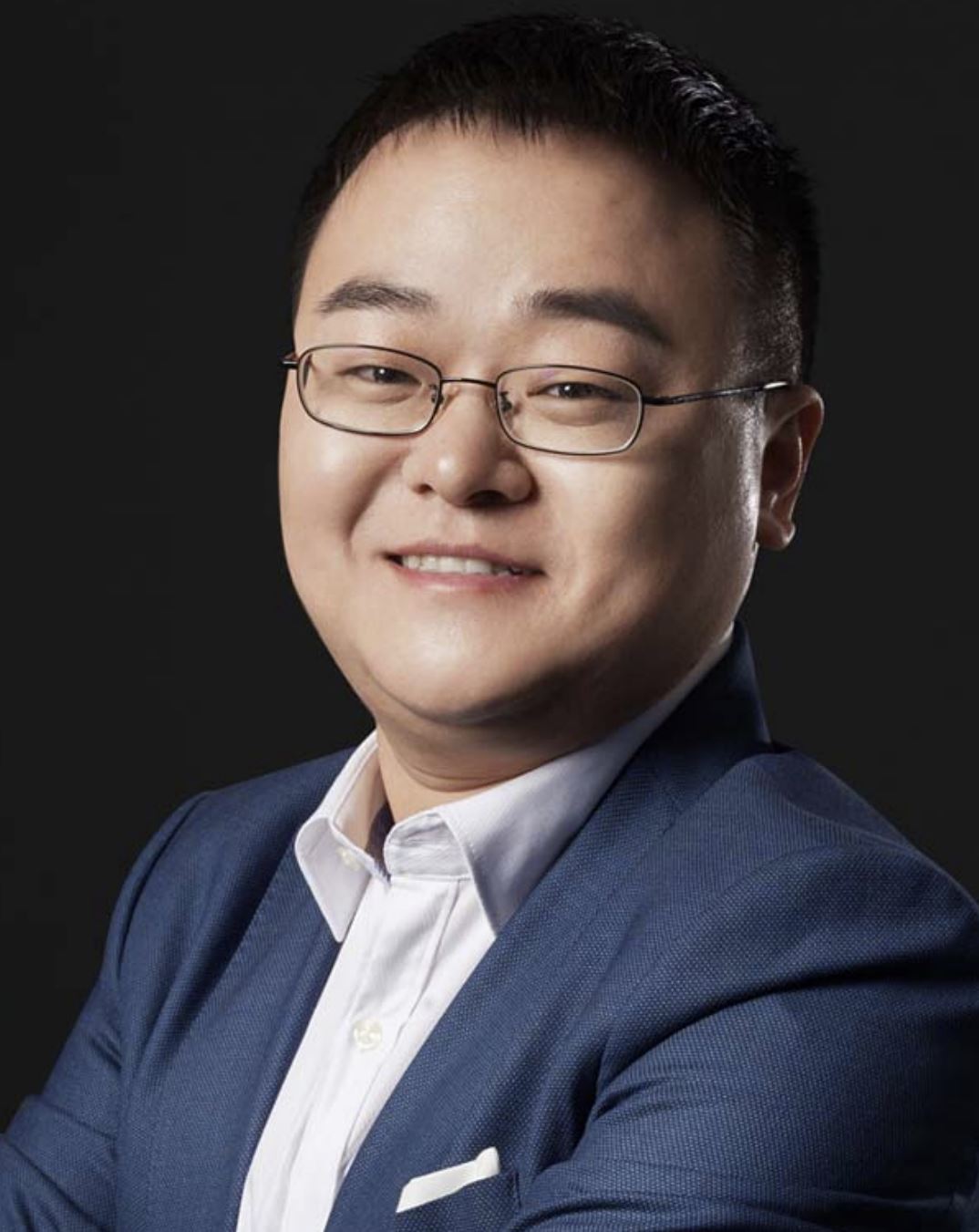}}]{Qiang Fu} received the B.S. and M.S. degrees from the University of Science and Technology of China, Hefei, China, in 2006 and 2009, respectively. 

He is the Director of the Game AI Center, Tencent AI Lab, Shenzhen, China. He has been dedicated to machine learning, data mining, and information retrieval for over a decade. His current research focus is game intelligence and its applications, leveraging deep learning, domain data analysis, reinforcement learning, and game theory.
\end{IEEEbiography}

\newpage
\begin{IEEEbiography}[{\includegraphics[width=1in,height=1.25in,clip,keepaspectratio]{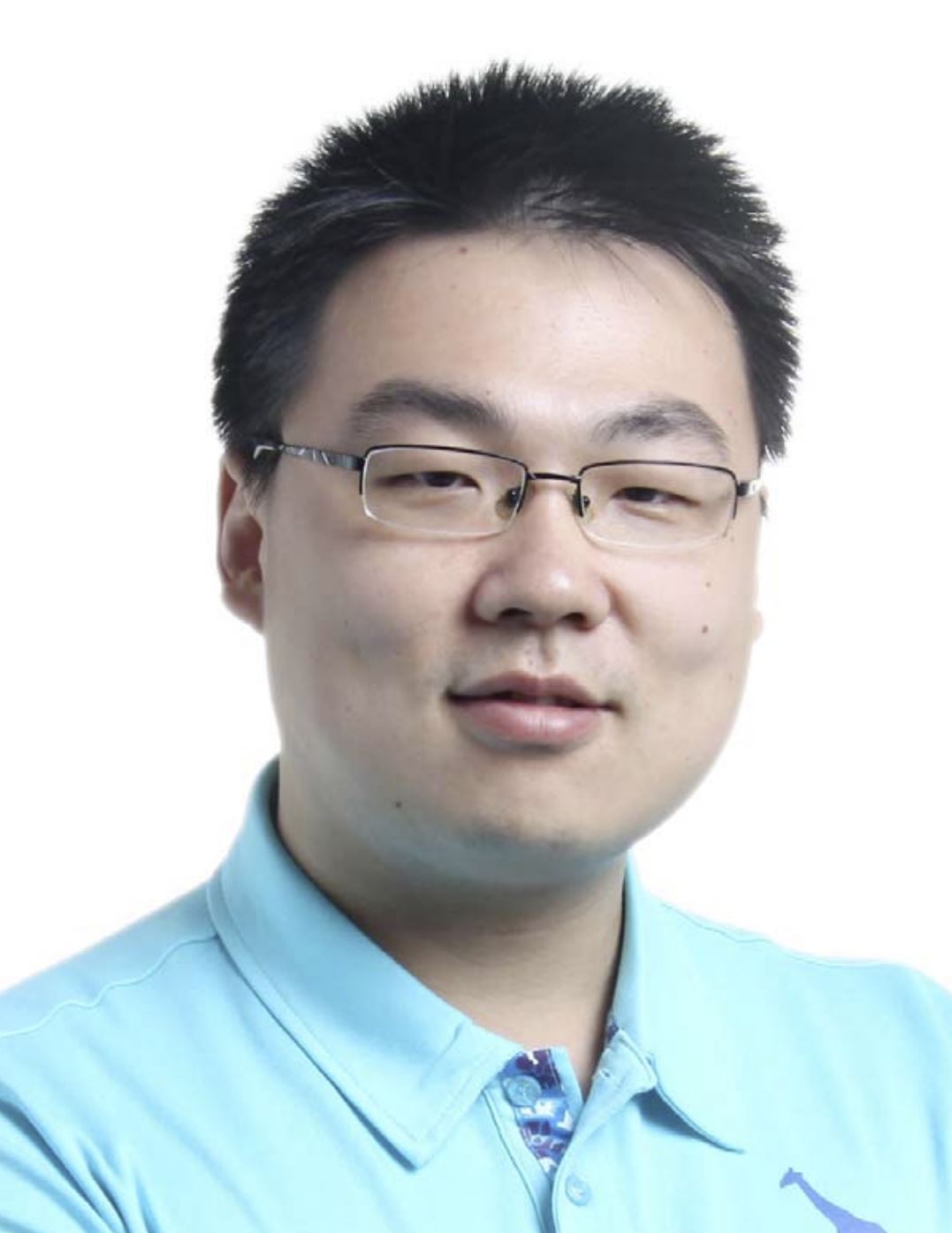}}]{Wei Yang} received the M.S. degree from the Huazhong University of Science and Technology, Wuhan, China, in 2007. 

He is currently the General Manager of the Tencent AI Lab, Shenzhen, China. He has pioneered many influential projects in Tencent in a wide range of domains, covering Game AI, Medical AI, search, data mining, large-scale learning systems, and so on.
\end{IEEEbiography}

\vspace{-1.1cm}
\begin{IEEEbiography}[{\includegraphics[width=1in,height=1.25in,clip,keepaspectratio]{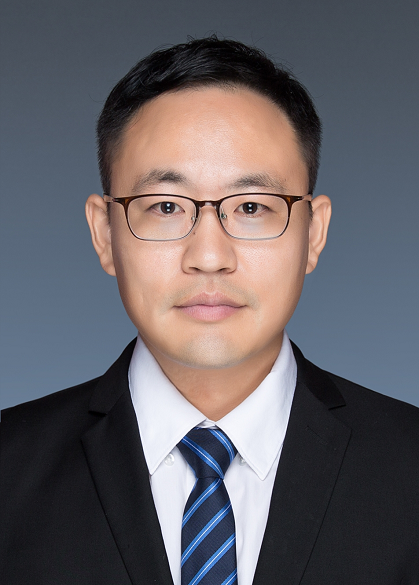}}]{Xueqian Wang} (Member, IEEE) received the M.Sc. and Ph.D. degrees in control science and engineering from the Harbin Institute of Technology (HIT),
Harbin, China, in 2005 and 2010, respectively. 

From June 2010 to February 2014, he was a Post-Doctoral Researcher with HIT. From March 2014 to November 2019, he was an Associate Professor with the Division of Informatics, Shenzhen International Graduate School, Tsinghua University, Shenzhen, China. He is currently a Professor and the Leader of the Center for Artificial Intelligence and Robotics, Shenzhen International Graduate School, Tsinghua University. His research interests include robot dynamics and control, teleoperation, intelligent decision-making and game playing, and fault diagnosis.
\end{IEEEbiography}

\vspace{-1.1cm}
\begin{IEEEbiography}[{\includegraphics[width=1in,height=1.25in,clip,keepaspectratio]{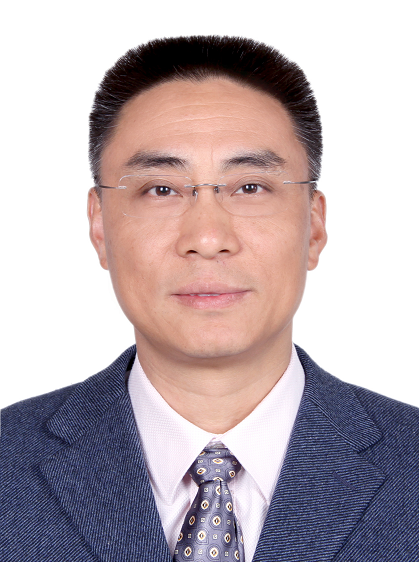}}]{Bin Liang}
(Senior Member, IEEE) received the B.Sc. and M.Sc. degrees in control engineering from the Honors College, Northwestern Polytechnical University, Xi’an, China, in 1989 and 1991, respectively, and the Ph.D. degree in control engineering from the Department of Precision Instrument, Tsinghua University, Beijing, China, in 1994.

From 1994 to 2003, he held his positions as a Post-Doctoral Researcher, an Associate Researcher, and a Researcher with the China Academy of Space Technology (CAST), Beijing. From 2003 to 2007, he held his positions as a Researcher and an Assistant Chief Engineer with the China Aerospace Science and Technology Corporation, Beijing. He is currently a Professor with the Research Center for Navigation and Control, Department of Automation, Tsinghua University. His research interests include modeling and control of intelligent robotic systems, teleoperation, and intelligent sensing technology.
\end{IEEEbiography}

\vspace{-1.1cm}
\begin{IEEEbiography}[{\includegraphics[width=1in,height=1.25in,clip,keepaspectratio]{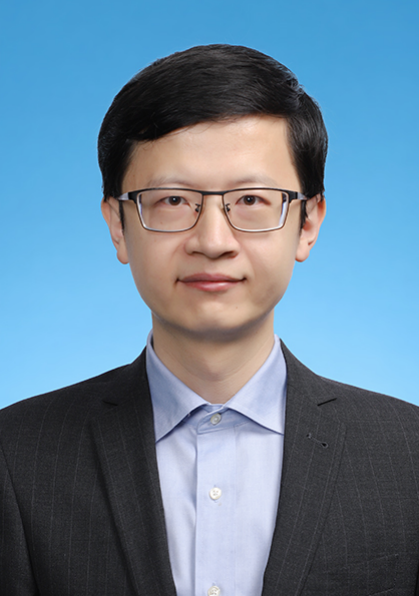}}]{Bo Yuan}
(Senior Member, IEEE) received the B.E. degree in computer science from the Nanjing University of Science and Technology, Nanjing, China, in 1998, and the M.Sc. and Ph.D. degrees in computer science from The University of Queensland (UQ), St Lucia, QLD, Australia, in 2002 and 2006, respectively.

From 2006 to 2007, he was a Research Officer on a project funded by the Australian Research Council, UQ. From 2007 to 2021, he was a faculty member (Lecturer: 2007-2009 and Associate Professor: 2009-2021) in the Division of Informatics, Tsinghua Shenzhen International Graduate School, P.R. China, and served as the Deputy Director of Office of Academic Affairs (2013-2020). In July 2021, he co-founded Shenzhen Wisdom \& Strategy Technology Co., Ltd., an innovative K-12 AI education solution provider. He has authored or coauthored more than 110 papers in refereed international conferences and journals. His research interests include data science, evolutionary computation, and reinforcement learning.
\end{IEEEbiography}

\vspace{-1.1cm}
\begin{IEEEbiography}[{\includegraphics[width=1in,height=1.25in,clip,keepaspectratio]{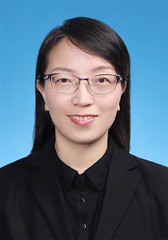}}]{Xiu Li}
(Member, IEEE) received her Ph.D. degree in computer integrated manufacturing from the Nanjing University of Aeronautics and Astronautics, Nanjing, China, in 2000. 

From 2003 to 2010, she was an Associate Professor with the Department of Automation, Tsinghua University, Beijing, China. From 2010 to 2016, she was an Associate Professor with the Division of Informatics, Shenzhen International Graduate School, Tsinghua University, Shenzhen, China. She is currently a Professor and the director of the Division of Informatics, Shenzhen International Graduate School, Tsinghua University. Her research interests include intelligent system, pattern recognition, and data mining.
\end{IEEEbiography}

\newpage

\appendices
\section{Illustration of Notations}
\label{appdix:notations_illustration}
 The notations and their detailed descriptions used in this article are summarized in Table \ref{table:notations}.

\begin{table}[h]
\centering
\setlength{\tabcolsep}{1.5mm}
\renewcommand\arraystretch{1.1}
\caption{Notations and their descriptions}
\label{table:notations}
\begin{tabular}{ll}
\toprule
Notation                        & Description                                                                    \\ \hline
$\mathcal{M}$                   & space of MDPs                                                                  \\
$\mathcal{D}$                   & dynamic environment over time in $\mathcal{M}$                                 \\
$M_t$                           & stationary task in the $t^{\rm th}$ time period                                \\
$x_t$                           & feature vector of $M_t$                                                        \\
$\theta_t$                      & weights of policy network in time period $t$                                   \\
$\pi_{\theta_t^{\ast}}$         & approximate optimal policy over $[M_1,M_2,\dots,M_t]$                             \\
$\pi_r$                         & uniform random policy                                                          \\
$z_t, z_t^\ast$                 & latent and assigned context label of task $M_t$                                \\
$z_{1:t}^\ast$                  & assigned context labels $(z_1,z_2,...,z_t)$ for $[M_1,M_2,\dots,M_t]$          \\
$\alpha$                        & concentration parameter in CRP                                                 \\
$m_{k}^{(t)}$                   & number of assignments to context $k$ up to the $t^{\rm th}$ time period        \\
$K_{t}$                         & number of instantiated contexts up to the $t^{\rm th}$ time period             \\
$\varphi_k^{(t)}$               & estimation of parameters of context $k$ in time period $t$                     \\
$\mu_k^{(t)}$                   & estimation of centroid vector of context $k$ in time period $t$                \\
$\theta_\mathcal{S}$            & weights of the shared representation layers in policy network                  \\
$\theta_{\mathcal{H},k}$        & weights of the $k^{\rm th}$ output head in policy network                      \\
$\beta$                         & learning rate for policy update                                                \\ 
$\lambda$                       & coefficient of distillation regularization                                     \\

\bottomrule
\end{tabular}
\end{table}

\section{Implementation Details}
\subsection{Environments}
\begin{figure}[!t]
  \centering
  \setlength{\abovecaptionskip}{-10pt}
  {\includegraphics[width=0.98\linewidth]{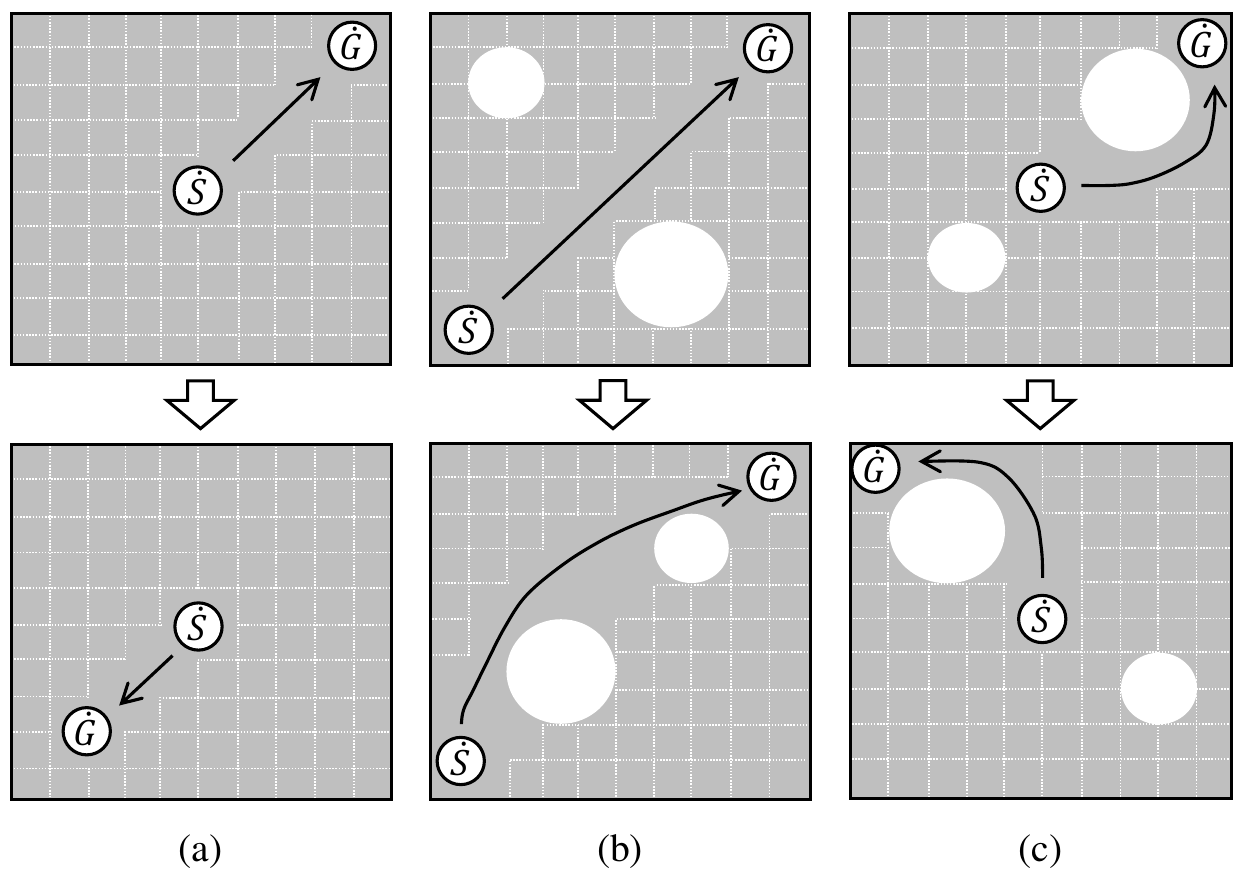}}
  \caption{Examples of three types of dynamic environments in the navigation tasks. $\dot{S}$ is the start point and $\dot{G}$ is the goal point. Puddles are shown in white. (a) Type I: the goal changes. (b) Type II: the puddles change. (c) Type III: both the goal and the puddles change.}
  \label{appdix:navigation}
\end{figure}

\begin{figure}[!t]
    \centering
    \setlength{\abovecaptionskip}{5pt}
    \subfigure[]
        {\includegraphics[width=0.310\linewidth]{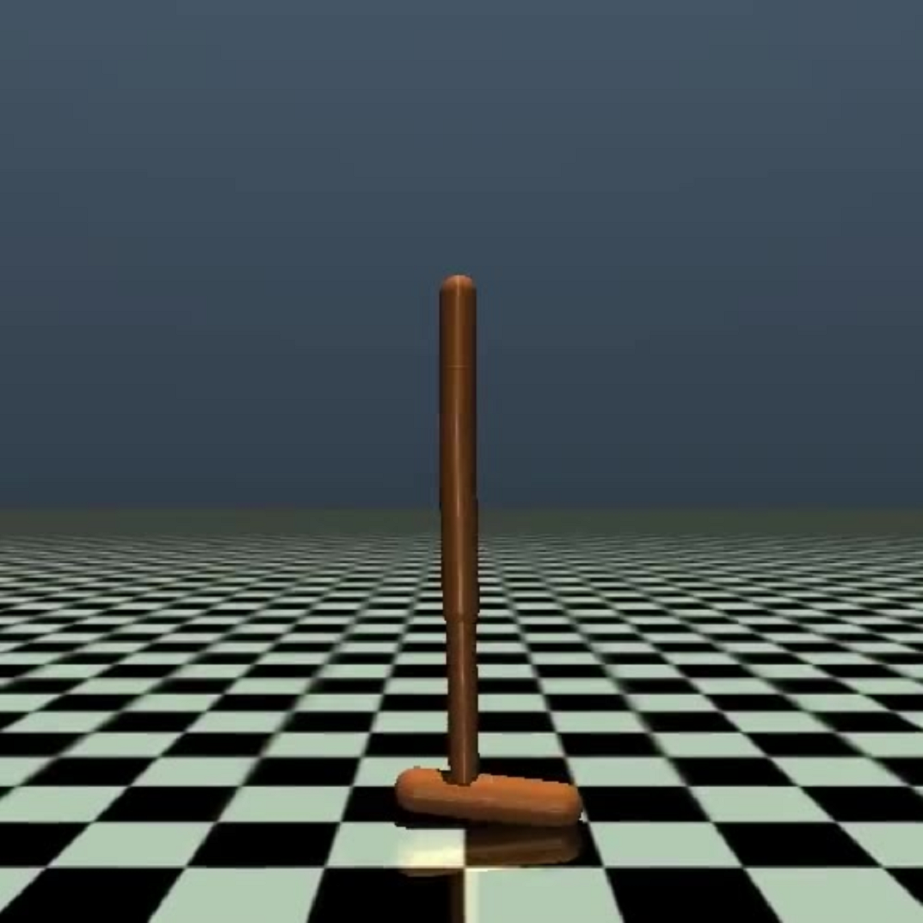}}
    \subfigure[]
        {\includegraphics[width=0.310\linewidth]{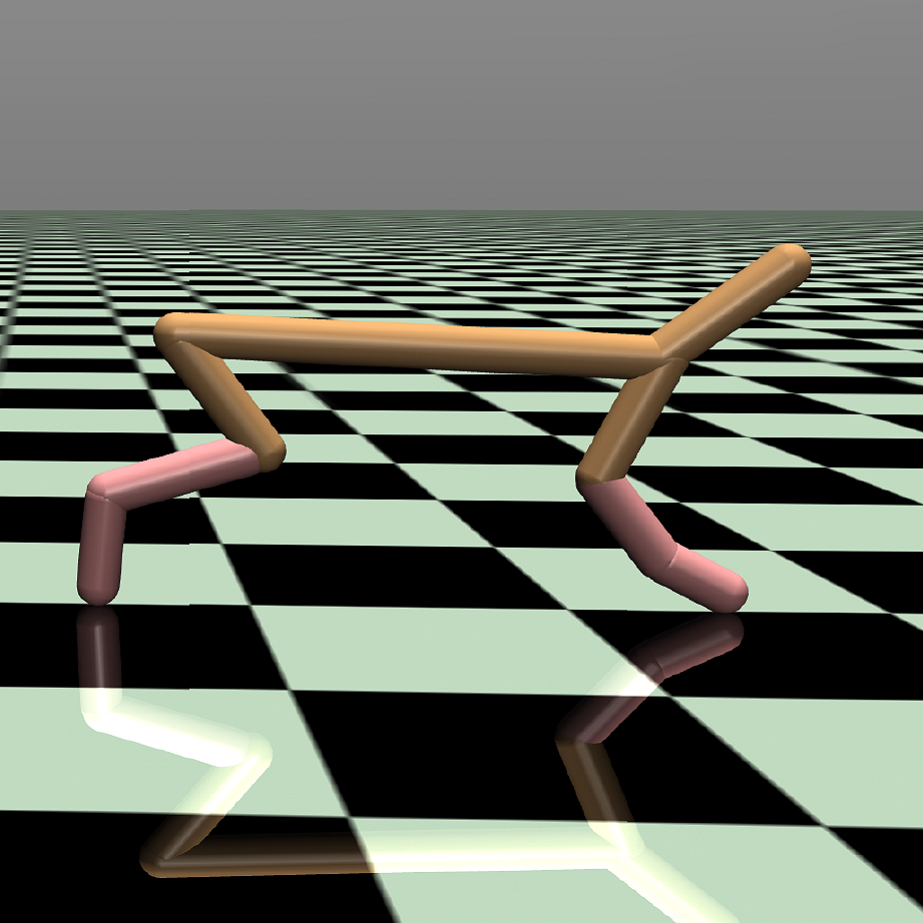}}
    \subfigure[]
        {\includegraphics[width=0.310\linewidth]{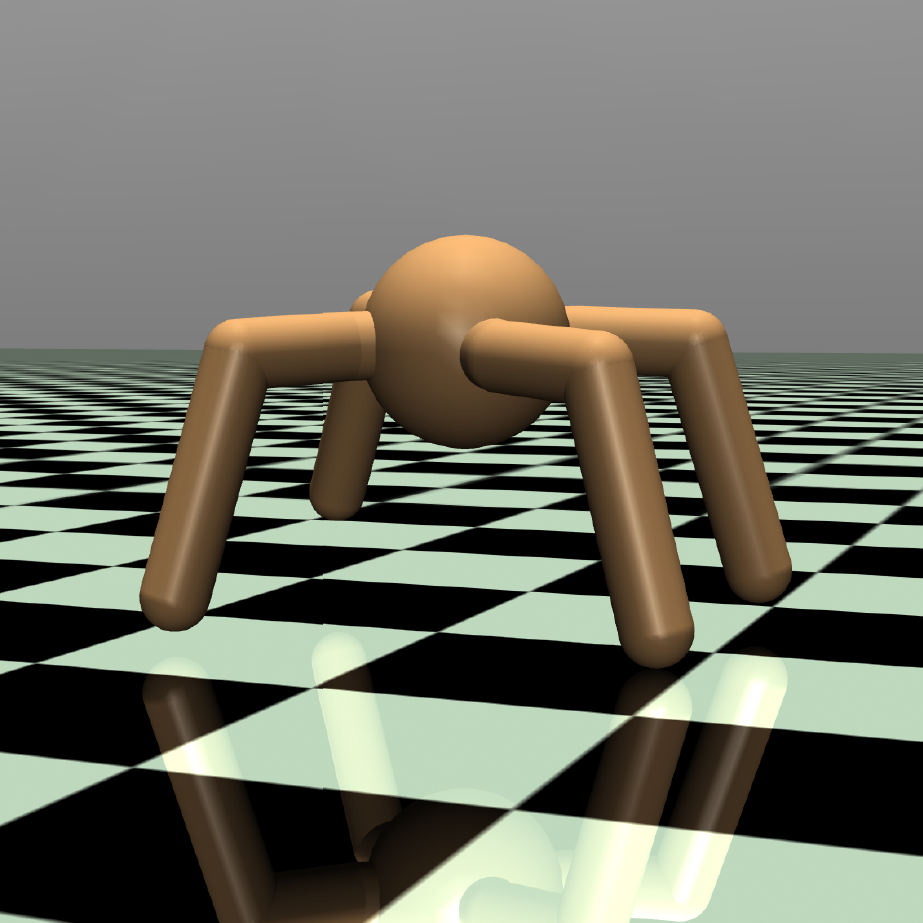}}
    \caption{Representative MuJoCo locomotion tasks with growing dimensions of state-action spaces including (a) Hopper, $|\mathcal{S}|=11$, $|\mathcal{A}|=3$, and $r=1-4\cdot|v_x-x_g|$. (b) HalfCheetah, $|\mathcal{S}|=20$, $|\mathcal{A}|=6$, and $r=-|v_x-x_g|$. (c) Ant, $|\mathcal{S}|=111$, $|\mathcal{A}|=8$, and $r=1-3\cdot|v_x-x_g|$. $v_x$ is the agent’s velocity in the positive x-direction and $v_g$ is the target velocity.}
    \label{appdix:mujoco}
\end{figure}

We present the illustrations of the Navigation and MuJoCo Locomotion tasks we used in our experiments in Fig. \ref{appdix:navigation} and Fig. \ref{appdix:mujoco}, respectively, to provide readers a more intuitively understanding of the benchmarks we utilized.

\subsection{Hyperparameters}
\label{appdix:hyperparas}
To ensure the fairness of comparison, our results compare the agents based on the underlying RL algorithm with the same hyperparameters. The choices for common key hyperparameters of REINFORCE and PPO we utilized are summarized in Table \ref{table:reinforce_parameter} and Table \ref{table:ppo_parameter}, respectively.

\begin{table}[!t]
\centering
\setlength{\tabcolsep}{7mm}
\renewcommand\arraystretch{1.2}
\caption{REINFORCE hyperparameters used in our experiments}
\label{table:reinforce_parameter}
\begin{tabular}{c|c}
\toprule
Hyperparameter             & Value               \\  \hline
Horizon                    & $100$               \\
SGD stepsize               & $2.0\times10^{-2}$    \\
Num. epochs                & $1$                 \\
Batch size                 & $16$                \\
Discount ($\gamma$)        & $0.95$              \\
\bottomrule
\end{tabular}
\end{table}

\begin{table}[!t]
\centering
\setlength{\tabcolsep}{5mm}
\renewcommand\arraystretch{1.2}
\caption{PPO hyperparameters used in our experiments}
\label{table:ppo_parameter}
\begin{tabular}{c|c}
\toprule
Hyperparameter                   & Value                 \\  \hline
Horizon                          & $100$                 \\
Adam stepsize                    & $1.0\times10^{-3}$    \\
Num. epochs                      & $5$                   \\
Batch size                       & $16$                  \\
Discount ($\gamma$)              & $0.95$                \\
Clipping parameter $\epsilon$    & $0.2$                 \\
\bottomrule
\end{tabular}
\end{table}

\begin{table*}[!t]
\centering
\setlength{\tabcolsep}{3.mm}
\renewcommand\arraystretch{1.19}
\caption{Numerical results in terms of the average forgetting performance over all sequential tasks during training in the\\ robot Navigation and MuJoCo Locomotion tasks (The top two performance are marked in boldface.)}
\label{table:Forgetting}
\begin{tabular}{c|c|c|c|c|c|c}
\toprule
Task       & Type I               & Type II             & Type III             & Hopper                & HalfCheetah           & Ant                   \\ \hline
Naive      & $26.58\pm18.89$      & $16.66\pm8.49$      & $29.61\pm7.37$       & $16.74\pm1.60$        & $23.47\pm8.90$        & $10.65\pm8.90$        \\
CRLUnsup   & $1.64\pm0.62$        & $21.77\pm6.37$      & $22.56\pm13.36$      & $28.69\pm23.49$       & $5.97\pm3.44$         & $-0.93\pm0.91$        \\
CDKD       & \bm{$-0.13\pm0.26$}  & \bm{$5.42\pm1.56$}  & \bm{$7.35\pm2.16$}   & \bm{$-2.95\pm3.26$}   & \bm{$-3.89\pm2.50$}   & \bm{$-13.25\pm3.48$}  \\
LLIRL      & $5.67\pm1.10$        & $17.61\pm1.98$      & $18.25\pm2.47$       & $0.36\pm5.43$         & $-3.03\pm1.30$        & $-10.89\pm3.79$       \\
DaCoRL     & \bm{$0.45\pm0.14$}   & \bm{$1.09\pm0.65$}  & \bm{$8.01\pm4.35$}   & \bm{$-1.61\pm0.77$}   & \bm{$-4.29\pm0.47$}   & \bm{$-11.33\pm3.75$}  \\
\bottomrule
\end{tabular}
\end{table*}

\begin{table}[!t]
\centering
\setlength{\tabcolsep}{2.2mm}
\renewcommand\arraystretch{1.2}
\caption{Numerical range of the concentration parameter $\alpha$ \\ related to different numbers of instantiated\\ contexts in the navigation tasks}
\label{table:concentration_parameter}
\begin{tabular}{c|c|c|c}
\toprule
$K_T$             & Type I                   & Type II                  & Type III               \\ \hline
1                 & $[0.000,0.005]$          & $[0.000,0.323]$          & $[0.000,0.011]$        \\                                   
2                 & $[0.006,0.222]$          & $[0.324,0.463]$          & $[0.012,0.133]$        \\
3                 & $[0.223,0.296]$          & $[0.464,0.619]$          & $[0.134,0.187]$        \\
4                 & $[0.297,0.632]$          & $[0.620,0.846]$          & $[0.188,0.818]$        \\
5                 & $[0.633,0.936]$          & $-$                      & $[0.819,0.931]$        \\
6                 & $[0.937,0.944]$          & $[0.847,0.890]$          & $[0.932,0.959]$        \\
7                 & $-$                      & $[0.891,0.997]$          & $-$                    \\
8                 & $[0.945,0.972]$          & $-$                      & $-$                    \\
\bottomrule
\end{tabular}
\end{table}

\begin{table}[!t]
\centering
\setlength{\tabcolsep}{1.2mm}
\renewcommand\arraystretch{1.1}
\caption{Statistics of agent-environment interaction episodes\\ in the training process of all methods}
\label{table:sample_efficiency}
\begin{tabular}{cccc}
\toprule
Methods      & Context Detection           & Policy Learning            & Total                        \\  \hline
Naive        & \multirow{2}{*}{$-$}        & \multirow{2}{*}{$800,000$}     & \multirow{2}{*}{$800,000$}     \\                                   
CRLUnsup     &                             &                            &                              \\  
\hline
CDKD         & \multirow{3}{*}{$800$}    & \multirow{3}{*}{$800,000$}     & \multirow{3}{*}{$800,800$}    \\
LLIRL        &                             &                            &                              \\
DaCoRL       &                             &                            &                              \\
\bottomrule
\end{tabular}
\end{table}

\section{Additional Experimental Results}
\subsection{Concentration Parameter Analysis}
\label{appdix:concentration_paras}
In DaCoRL, we implicitly control the number of contexts that need to be instantiated by adjusting the concentration parameter $\alpha$ in the CRP prior distribution. 
In order to investigate the effect of $\alpha$ on the results of incremental clustering and better grasp the characteristics of this parameter tuning, we vary the concentration parameter $\alpha$ in the CRP to obtain different numbers of instantiated contexts $K_T$ in the three types of navigation tasks. 
The results of numerical range of $\alpha$ related to specific $K_T$ are shown in Table \ref{table:concentration_parameter}. 
It is clearly to see that $\alpha$ is quite insensitive and all $\alpha$ values within the specific continuous interval can result in the same context instantiation results. For example, in our experiments, all values of $\alpha$ in $[0.633, 0.936]$ for Type I, in $[0.847, 0.890]$ for Type II and in $[0.188, 0.818]$ for Type III can instantiate the same number of contexts as expected in Fig. \ref{fig:navi_performance}, i.e., $K_T=5$ for Type I, $K_T=6$ for Type II and $K_T=4$ for Type III.
These results demonstrate that compared with predicting the number of clusters in advance without any prior knowledge of the dynamic environment, it is more technical and practical to automatically group the cluster patterns by tuning the concentration parameter $\alpha$.

\subsection{Sample Efficiency}
\label{appdix:sample_efficiency}
In our experiments, the algorithm performs a policy iteration once every 16 episodes of agent-environment interaction. We train the model for 1k policy iterations on the corresponding task during each time period.
Before policy learning starts in each time period, methods with context detection phase (i.e., CDKD, LLIRL and DaCoRL) are required to explore the current environment using a uniform policy and collect extra samples for context inference.
Table \ref{table:sample_efficiency} records the number of agent-environment interaction episodes during training on a sequential tasks ($T=50$) for all methods. The results show that, compared with the order of magnitude of interactions between the agents and the environments throughout the process of policy learning, the sample size needed for context detection is extremely modest and trivial. Meanwhile, the performance of the approaches with context detection module is significantly better than that of the baselines without this module according to the extensive experimental results in the main body of this article.

\subsection{Forgetting}
\label{appdix:forgetting}
In order to study the protection of our proposed method to the learned knowledge, we calculate the average forgetting (over $T=50$ sequential tasks) of all methods during training process and the results are summarized in Table \ref{table:Forgetting}.
Here, the positive values indicate that the agent trained by the associated methods forget a certain degree of learned knowledge during their learning process, while the negative values indicate positive backward transfer. 
We put the top two performance in bold for each dynamic environment, from which we can conclude that:
\begin{itemize}
    \item[1)] Since Naive learns the sequential tasks with a single model and without taking into account any mitigation of catastrophic forgetting, it exhibits the greatest degree of forgetting of the learned knowledge throughout the entire training process in all tasks. This is also the primary cause of the extensive results in our experiments showing the worst performance of this method.  
    \item[2)] Although CRLUnsup uses the EWC mechanism to guard the important weights of the previously learned tasks to prevent forgetting, it can be problematic to track long- and short-term average cumulative rewards to detect environmental changes when the initial cumulative reward for the agent on the new task does not fall off noticeably in comparison to that following training on the preceding task. As a result, depending in the task, the forgetting metric of CRLUnsup varies greatly.
    \item[3)] As a representative method of using multiple separate context-specific neural networks for fast adaptation, LLIRL coincidentally avoids the interference among tasks in different contexts and the resulting forgetting. Nevertheless, interference still persists among tasks within the same context because no any mitigation methods are applied. Therefore, the forgetting of LLIRL is generally lower than the above two approaches, but it still remains a high level in all three types of navigation tasks.
    \item[4)] By contrast, CDKD and DaCoRL can effectively alleviate the interference among tasks not only in different contexts but also within the same context by utilizing knowledge distillation technique and multi-head neural network design. Their performance in reducing forgetting on all tasks is either best or second best, which well demonstrates the efficiency of applying this technique and network architecture for continuous learning. 
\end{itemize}

\end{document}